\documentclass[twoside,11pt]{article}

\usepackage{amsthm} %

\usepackage[preprint]{jmlr2e}

\usepackage{amsmath}
\usepackage{amsfonts}
\usepackage{bm}
\usepackage{placeins}
\usepackage{adjustbox}
\usepackage{booktabs}
\usepackage{graphicx}
\usepackage{subcaption}
\usepackage{xcolor}

\theoremstyle{definition}
\newtheorem{definitiontwo}{Definition}[section]
\newtheorem{theoremtwo}{Theorem}

\newcommand{\norm}[1]{\left\lVert#1\right\rVert}
\newcommand{\X}{\mathcal{X}}
\newcommand{\Y}{\mathcal{Y}}
\renewcommand{\H}{\mathcal{H}}
\newcommand{\E}{\mathbf{E}}
\newcommand{\R}{\mathbb{R}}

\newcommand{\F}{\mathcal{F}}

\newcommand{\Div}{\text{Div}}
\renewcommand{\H}{\mathcal{H}}
\newcommand{\MMD}{\text{MMD}}

\newcommand{\x}{\bm{x}}
\newcommand{\y}{\bm{y}}
\newcommand{\Test}{\textcolor{green!80!black}{Test}}
\newcommand{\Vae}{\textcolor{red!80!black}{Vae}}
\newcommand{\Gan}{\textcolor{orange!100!black}{Gan}}

\DeclareMathOperator*{\argmax}{\arg\max}

\definecolor{mygray}{rgb}{0.7, 0.7, 0.7}

\newcommand{\innerProd}[2]{\left\langle #1 , #2 \right\rangle}

\hypersetup{ %
	pdftitle={Parametric Adversarial Divergences are Good Losses for Generative Modeling },
	pdfkeywords={},
	pdfborder=0 0 0,
	pdfpagemode=UseNone,
	colorlinks=true,
	linkcolor=blue, %
	citecolor=blue, %
	filecolor=blue, %
	urlcolor=blue, %
	pdfview=FitH,
	pdfauthor={Gabriel Huang, Hugo Berard, Ahmed Touati, Gauthier Gidel, Pascal Vincent, Simon Lacoste-Julien},
}

\jmlrheading{1}{2021}{1-48}{4/00}{10/00}{meila00a}{Gabriel Huang, Hugo Berard, Ahmed Touati, Gauthier Gidel, Pascal Vincent, Simon Lacoste-Julien}

\ShortHeadings{Parametric Adversarial Divergences}{Huang, Berard, Touati, Gidel, Vincent and Lacoste-Julien}
\firstpageno{1}

\begin{document}

\title{Parametric Adversarial Divergences are \\Good Losses for Generative Modeling }

\author{\name Gabriel Huang \textsuperscript{1}\\
Hugo Berard \textsuperscript{1,2}\\
Ahmed Touati \textsuperscript{1,2}\\
Gauthier Gidel \textsuperscript{1}\\
Pascal Vincent \textsuperscript{1,2,3}\\
Simon Lacoste-Julien \textsuperscript{1,3} \email \{firstname.lastname\}@umontreal.ca \\
       \addr 
\textsuperscript{1}Mila \& University of Montreal, Canada\\
\textsuperscript{2}Facebook AI Research, 
       Canada\\
\textsuperscript{3}Canada CIFAR AI Chair, 
       Canada\\}

\editor{Francis Bach, David Blei and Bernhard Sch{\"o}lkopf}

    \maketitle

\begin{abstract}%
\emph{Parametric adversarial 	divergences}\footnote{We give a formal definition in this paper.}, which are a generalization of the losses used to train generative adversarial networks (GANs), have often been described as being approximations of their nonparametric counterparts, such as the Jensen-Shannon divergence, which can be derived under the so-called \textit{optimal discriminator} assumption. 
In this position paper, we argue that despite being ``non-optimal'', parametric divergences have distinct properties from their nonparametric counterparts which can make them more suitable for learning high-dimensional distributions.
A key property is that parametric divergences are only sensitive to certain aspects/moments of the distribution, which depend on the architecture of the discriminator and the loss it was trained with. 
In contrast, nonparametric divergences such as the Kullback-Leibler divergence are sensitive to moments ignored by the discriminator, but they do not necessarily correlate with sample quality~\citep{theis2015note}.
Similarly, we show that mutual information can lead to unintuitive interpretations, and explore more intuitive alternatives based on parametric divergences.
We conclude that parametric divergences are a flexible framework for defining statistical quantities relevant to a specific modeling task.

\end{abstract}


\section{Introduction}

In traditional statistics, generative modeling is formulated as density estimation. The learning objective and evaluation metric are usually the expected negative-log-likelihood. While maximizing the log-likelihood, or equivalently, minimizing the KL-divergence, works fine for modeling low-dimensional data, there are a number of issues that arise when modeling high-dimensional data, such as images. 
Maybe the most important issue is the lack of guarantees that log-likelihood is a good proxy for sample quality. 
This is obviously a problem, because if the goal is to generate realistic data, then the evaluation objective \textit{should} align with sample quality.
For instance, \citet{theis2015note} exhibit generative models with high log-likelihood which produce low-quality images, and models with poor log-likelihood which produce high-quality images. In some cases, they show that the log-likelihood can even be \textit{hacked} to be arbitrarily high, even on test data, without improving the sample quality at all. Another practical issue with f-divergences, a generalization of the KL, is they are either not defined or uninformative whenever the distributions are too far apart, even when tricks such as smoothing are used~\citep{arjovsky2017wasserstein}.

To address these shortcomings, we look past maximum-likelihood and classic divergences to define better objectives. 
But first, we need to take a step back and explicit our \textit{final task}, or end goal.
Assuming that our end goal is to generate \textit{realistic} and \textit{diverse} samples,  how can we formalize such a subjective and ill-defined final task into a \textit{task loss},
a rigorous mathematical objective which can be evaluated and used to derive training (e.g. surrogate) losses? %
When it comes to defining relevant task losses, it is worthwhile to consider how people choose task losses in structured prediction, for which the label space is often combinatorially large (e.g. sequences of words).
For instance, machine translation systems are commonly evaluated using the BLEU-4 metric~\citep{papineni2002bleu}, which essentially counts how many words the predicted and ground truth sentences have in common. 
Although the BLEU score is only an imperfect approximation of the final task, it is nevertheless more informative than ``hard'' losses such as the $0-1$ loss, which give no training signal unless the predicted label matches exactly the ground truth.

Unfortunately, in generative modeling, it is not as obvious how to define a task loss that correlates well with diversity and sample quality. Nevertheless, we argue that the \textit{adversarial framework}, introduced in the context of generative adversarial networks or GANs~\citep{goodfellow2014generative}, provides an interesting way to define meaningful and practical task losses for generative modeling. For that purpose, we adopt the view\footnote{We focus in this paper on the divergence minimization perspective of GANs. There are other views, such as those based on game theory \citep{arora2017generalization,fedus2017many}, ratio matching and moment matching \citep{mohamed2016learning}, but these are outside the scope of this paper.} that training a GAN can be seen as training an implicit generator to minimize a special type of task loss, which we call \textbf{parametric (adversarial) divergence}:
\begin{align} \label{eq:nndiv}
\Div(p||q_\theta) \widehat =& \sup_{\phi \in \Phi} \E_{(\x,\x')\sim p\otimes {q_\theta}}[\Delta(f_\phi(\x), f_\phi(\x'))]
\end{align}
where $p$ is the distribution to learn and $q_\theta$ is the distribution defined by the implicit generator.  The expectation is maximized over a parametrized class of functions $\{f_\phi:\X\rightarrow\R^{d'} \;;\; \phi\in\Phi\}$ which are usually neural networks with a fixed architecture. Those functions are called \textbf{discriminators} in the GAN framework~\citep{goodfellow2014generative}. The constraints $\Phi$ and the formulation $\Delta: \R^{d'} \times \R^{d'}\rightarrow\R$ determine properties of the resulting divergence (see Section~\ref{sub:adversarial_and_traditional_divergences} for concrete examples).

The main contribution of this paper is to show that parametric adversarial divergences can have very different properties from their \emph{nonparametric} counterparts (where the function class is infinite dimensional). Instead of viewing them as imperfect estimators, we argue that parametric divergences are actually better approximators of ``generating realistic samples'' than likelihood-based objectives. 
To this end, we start by expliciting the difference between final task and task loss (Section~\ref{sec:priortask}). Then, we show that unlike many nonparametric divergences, parametric divergences offer favorable sample complexity while retaining the flexibility to adapt to the final task (Section~\ref{sec:samplecomplexity}). 
In particular, we show on a toy problem how to tune a parametric divergence in order to enforce properties of interest (Section~\ref{sec:reflecttask}). 
In practice, combining divergences with specific generators can lead to side-effects, which we discuss in Section~\ref{sec:interactions}.
Finally, we investigate how to use parametric divergences to define more intuitive notions of mutual information (Section~\ref{sec:pmi}).

\section{Background}

We briefly introduce the structured prediction framework because we will make links between the losses in structured prediction and generative modeling in Section~\ref{sec:hardsoftstructuredpred}.  We introduce the variational formulation which allows us to consider and compare parametric adversarial divergences and traditional divergences as special cases of adversarial divergences.

\subsection{Structured Prediction}

The goal of structured prediction is to learn a function $h_\theta: \X\rightarrow\Y$ which predicts a structured output $\y$ from an input $\x$. 
Examples of structured outputs include parse-trees, sequences of symbols, alignments between sequences, 3D shapes, and segmentation maps.
The key difficulty is that $\Y$ usually has size exponential in the dimension of the input (e.g. for sequence-to-sequence prediction, $\Y$ could be all the sequences of symbols with a given length). Being able to handle this exponentially large set of possible outputs is one of the key challenges in structured prediction. Traditional multi-class classification methods are unsuitable for these problems in general.
Standard practice in structured prediction~\citep{Taskar2003,collins2002discriminative,pires2013cost} is to consider predictors based on score functions
	$h_\theta(\x) \widehat= \argmax_{\y' \in \Y} s_\theta(\x,\y')$,
where $s_\theta: \X\times\Y \to \R$, called the \textbf{score/energy function}~\citep{lecun2006tutorial}, assigns a score to each possible label $\y$ for an input $\x$. Typically, as in structured SVMs~\citep{Taskar2003}, the score function is linear: $s_{\theta}(\x,\y) = \innerProd{\theta}{g(\x,\y)}$, where $g(\cdot)$ is a predefined feature map. Alternatively, the score function could also be a learned neural network~\citep{belanger2016structured}.

In order to evaluate the predictions objectively, we need to define a \textbf{task-specific} structured loss $\ell(\y',\y \,; \x)$ which expresses the cost of predicting $\y'$ for $\x$ when the ground truth is~$\y$. We discuss the relation between the loss function and the actual final task when we review statistical decision theory in Section~\ref{sec:welldefinedtask} and~\ref{sec:ParallelsGenerative}.
The goal is then to find a parameter $\theta$ which minimizes the generalization error
\begin{equation}\label{eq:generalization_risk}
	\min_{\theta\in\Theta} \mathbf{E}_{(\x,\y) \sim p} \left[ \ell(h_{\theta}(\x),\y,\x) \right]
\end{equation}
or, in practice, an empirical estimation of it based on an average over a finite sample from $p$. Directly minimizing this is often intractable, even in simple cases, e.g. when the structured loss~$\ell$ is the 0-1 loss~\citep{arora1993hardness}. Instead, the usual practice is to minimize a surrogate loss $\mathbf{E}_{(\x,\y) \sim p} \left[\mathcal L(s_{\theta}(\x, \y), \y,\x) \right]$ \citep{bartlett2006convexity} which has nicer properties, such as sub-differentiability or convexity, to get a tractable optimization problem.
The surrogate loss is said to be consistent~\citep{osokin2017structured} when its minimizer is also a minimizer of the task loss.

\subsection{Parametric and Nonparametric Adversarial Divergences} %
\label{sub:adversarial_and_traditional_divergences}

The focus of this paper is to analyze whether parametric adversarial divergences are good candidates for generative modeling. In particular, we analyze them relatively to nonparametric divergences. Therefore, we first unify them with a formalism similar to~\citet{sriperumbudur2012empirical, liu2017approximation}. We define \textbf{adversarial divergences} using the variational formulation:
\begin{definitiontwo}[Adversarial Divergence]
We denote adversarial divergences functions which can be written with the following form:
\begin{align}\label{eq:unify2}
\Div(p||q_\theta) \widehat =& \sup_{f\in\F} \E_{(\x,\x')\sim p\otimes {q_\theta}}[\Delta(f(\x), f(\x'))]
\end{align}
where we refer to $f:\X\rightarrow\R^{d'}\in \mathcal F$ as the discriminator, and $\Delta: \R^{d'} \times \R^{d'}\rightarrow\R$ determines properties of the resulting divergence.
\end{definitiontwo}

\begin{definitiontwo}[Parametric Divergence]
In particular, when the discriminator space $\F$ is \textit{parametric}, such as the set of neural networks with a given architecture, the adversarial divergence is called a parametric (adversarial) divergence.\footnote{Usually, $\F$ is a class of neural networks with fixed architecture. In that case, $\Div(p||q_\theta)$ has been called a \textbf{neural divergence} in~\citet{arora2017generalization}. We will use the slightly more generic \textbf{parametric divergence} in our work.} 
\end{definitiontwo}

For appropriate choices of discriminator class $\F$ and function $\Delta$, we can recover many usual divergences, including f-divergences (such as Kullback-Leibler) and integral probability metrics (such as Wasserstein distances and Maximum Mean discrepancy). For instance,
\begin{itemize}
\item $\psi$-divergences with generator function $\psi$ (which we call f-divergences) can be written in dual form~\citep{nowozin2016f}\footnote{The standard form is $\E_{\x\sim q_\theta}[\psi(\frac{p(x)}{q_\theta(x)})]$.}
\begin{equation}\label{eq:fdiv}
\Div_\psi(p||q_\theta)  \widehat  =  \!\!\!\sup_{\substack{f: \X\rightarrow\R}} \E_{\x\sim p}[f(\x)] - \E_{\x'\sim q_\theta}[\psi^*(f(\x'))]
\end{equation}
where $\psi^*$ is the convex conjugate. Depending on $\psi$, one can obtain any $\psi$-divergence such as the (reverse) Kullback-Leibler, the Jensen-Shannon, the Total Variation, the Chi-Squared.\footnote{For instance the Kullback-Leibler $\E_{x\sim p}[\log \frac{p(x)}{q_\theta(x)}]$ has the dual form $\sup_{\substack{f: \X\rightarrow\R}} \E_{\x\sim p}[f(\x)] - \E_{\x'\sim q_\theta}[\exp(f(\x')-1)]$.
Some $\psi$ require additional constraints, such as $||f||_\infty \leq 1$ for the Total Variation.}
\item Wasserstein-1 distance induced by an arbitrary norm $\norm{\cdot}$ and its corresponding dual norm $\|\cdot\|^*$ ~\citep{sriperumbudur2012empirical}:
\begin{equation}\label{eq:w}
W(p||q_\theta) \, \widehat = \hspace{-3mm} \sup_{\substack{f: \X\rightarrow\R\\ \forall \x\in\X,\\ ||f'(\x)||^* \leq 1}} \E_{\x\sim p}[f(\x)] - \E_{\x'\sim q_\theta}[f(\x')]
\end{equation}
which can be interpreted as the cost to transport all probability mass of $p$ into $q$, where $\norm{\x-\x'}$ is the unit cost of transporting $\x$ to $\x'$.
\item Maximum Mean Discrepancy~\citep{gretton2012kernel}:
\begin{equation}\label{eq:mmd}
\MMD(p||q_\theta) \widehat = \!\!\!\sup_{\substack{f\in\H\\ \|f\|_{\H} \leq 1}} \E_{\x\sim p}[f(\x)] - \E_{\x'\sim q_\theta}[f(\x')]
\end{equation}
where $(\H, K)$ is a Reproducing Kernel Hilbert Space induced by a Kernel $K(\x,\x')$ on $\X$ with the associated norm $\|\cdot\|_{\H}$. The MMD has many interpretations in terms of moment-matching~\citep{li2017mmd}.
\end{itemize}

Most nonparametric divergences can be made parametric by replacing $\F$ with neural networks: examples are the \textbf{parametric Jensen-Shannon}, which is the standard mini-max GAN objective~\citep{goodfellow2014generative} and the \textbf{parametric Wasserstein} which is the WGAN objective~\cite{arjovsky2017wasserstein} in essence, modulo some technical tricks.\footnote{There are subtleties in the way the Lipschitz constraint is enforced. More details in~\citet{petzka2017regularization}.} See~\citet{liu2017approximation} for interpretations and a review and interpretation of other divergences like the Wasserstein with entropic smoothing~\citep{aude2016stochastic}, energy-based distances~\citep{li2017mmd} which can be seen as adversarial MMD, and the WGAN-GP~\citep{gulrajani2017improved} objective.

We deliberately chose a somewhat ambiguous terminology -- nonparametric v.s. parametric -- not to imply a clear-cut distinction between the two (as e.g. neural networks can be made to become universal function approximators as we increase their size), but to imply a continuum from least restricted to more restricted function families where the latter are typically expressed through an explicit parametrization.

Under the formalism~\eqref{eq:unify2}, one could argue that parametric divergences are simply estimators --in fact lower-bounds-- of their nonparametric counterparts. Our opinion is that parametric divergences are not merely convenient estimators, but can actually be \textit{much better} objectives for generative modeling than nonparametric divergences. We will give practical arguments and experiments to support this statement in the rest of this paper.

\section{Divergences as Task Losses\label{sec:priortask}}

While tasks such as binary and multiclass classification are straightforward to evaluate using classification accuracy (also known as the 0-1 loss), tasks with more structured outputs often require more complex evaluation. 
For instance, machine translation is often evaluated using the BLEU metric~\citep{papineni2002bleu}, text summarization using the ROUGE-L metric~\citep{lin2004rouge}, object detection using mean average precision~\citep{ren2015faster}, semantic segmentation the mean intersection-over-union~\citep{chen2017rethinking}, while several metrics such as Multiple Object Tracking Accuracy (MOTA) and Multiple Object Tracking Precision (MOTP) are commonly used for evaluating video tracking~\citep{ciaparrone2020deep}.
However, optimizing these metrics is not an end in itself, rather the hope is that these metrics are good proxies for the final task (producing an accurate translation, a concise summary, relevant bounding boxes, precise video tracking).

Statistical decision theory is a general framework for modeling the task of acting (in our case, learning a model) under uncertainty (which can come from sampling noise). In particular, ill-defined tasks can be formalized as the minimization of an evaluation metric, called the task loss.

In this work, we propose to consider \textit{parametric divergences} as the task losses for approximating the ill-defined task of \textit{generating realistic samples}.
We start by introducing statistical decision theory and discuss desirable properties for task losses (Section~\ref{sec:welldefinedtask}). 
Then, we unify structured prediction and generative modeling under the statistical decision theory framework (Section~\ref{sec:ParallelsGenerative}). 
We point to results that hard task losses such as the 0-1 loss can make learning exponentially slower than using softer losses such as the Hamming loss (Section~\ref{sec:hardsoftstructuredpred}).
Those results suggest that similarly for generative modeling,  it might be beneficial to use ``softer'' parametric divergences instead of ``harder'' nonparametric divergences such as the Kullback-Leibler divergence.

%

\subsection{Formalizing Final Tasks with Statistical Decision Theory\label{sec:welldefinedtask}}

Statistical decision theory is the standard framework presented in statistic textbooks for modeling and evaluating the task of acting under uncertainty~\citep{bickel2015mathematical}. In our case, \textit{acting} means learning a model such as a classifier or a generative model from data, and the \textit{uncertainty} comes from the fact that we can only access finite samples from the true distribution.
Statistical decision theory allows us to formalize ill-defined tasks as the minimization of a clearly-defined evaluation metric, which is called the (statistical) \textit{task loss}. The task loss needs to be mathematically well-defined, %
and cannot, for instance, require human evaluation in the loop, which would make the metric subjective and ill-defined.

\paragraph{Notation.} We denote $p\in\mathcal P$ the unknown state of the process (distribution) we want to model, $a\in\mathcal A$ the action,  and $L_p(a)$ the task loss.
In machine learning, the unknown state is typically a probability distribution $p(\x)$ or $p(\x,\y)$ which we can only access indirectly through a finite training set $D_{train}$ sampled i.i.d.\ from $p$, and the learner ``plays an action'' such as choosing the classifier or generative model which minimizes the negative log-likelihood of the training set.

\begin{figure}
	\centering
	\begin{minipage}{0.6\linewidth}
		\includegraphics[width=\linewidth]{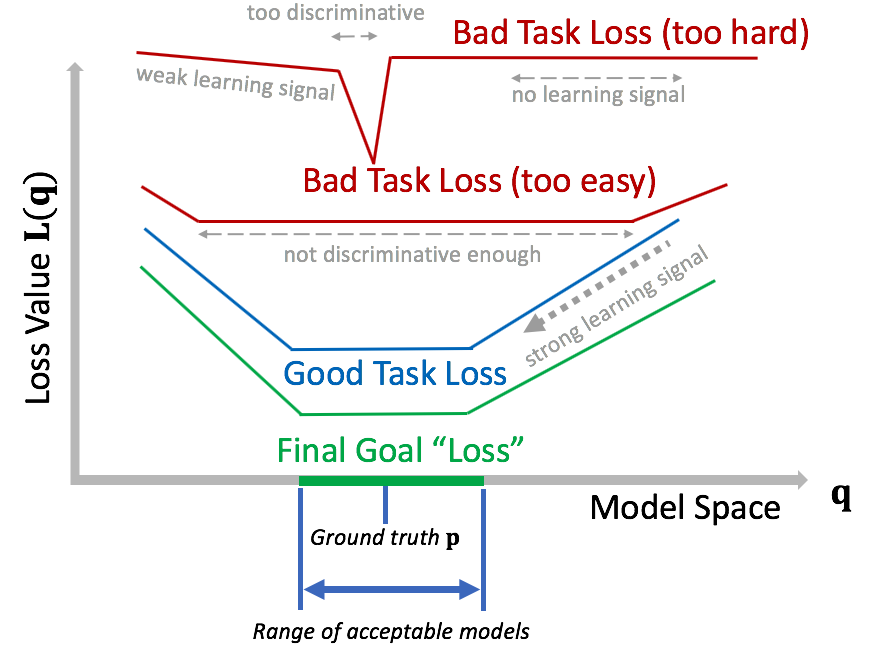}
	\end{minipage}
	\begin{minipage}{0.39\linewidth}
		\caption{\small
			Our goal is to solve the final task, represented by the \textcolor{green!60!black}{green} ``loss''. The first task loss (upper \textcolor{red}{red}) is not a good loss because it is \textbf{too discriminative} and rejects many acceptable models.
			The second task loss (lower \textcolor{red}{red}) is not good either because it is \textbf{not discriminative} enough and accepts incorrect models.
			The third task loss (\textcolor{blue}{blue}) is a good task loss because it ``tracks'' the final goal ``loss'' well.
			\label{fig:goodandbad}}
	\end{minipage}
\end{figure}

\paragraph{Designing task losses.} What does the ideal task loss look like?
Consider the example in Figure~\ref{fig:goodandbad}.
Our goal is to solve the final task,
which is illustrated by the \textcolor{green}{green} final goal ``loss''.
Such a ``loss'' might not be
expressible as a mathematical loss, for instance it could be a score based on human evaluation or perception.
Therefore, we need to approximate the final goal with a task loss $L_p(q_\theta)$, a proper mathematical objective from which we can derive an optimization problem. 
The task loss is formed with respect to a ground truth $p$ which typically corresponds to the target distribution for generative modeling, and to the ground-truth labels for structured prediction. 
The first task loss (topmost in \textcolor{red}{red}) is not a good loss because it is too ``hard'' (discriminative). It only accepts the ground-truth $p$ as a good model even though there exist a range of models with equivalent final goal loss. As a consequence, these losses tend to provide little to no learning signal since most models are ruled out as equally bad and the task loss ``saturates'' without providing adequate search direction. The second task loss (second in \textcolor{red}{red}) is not good either because it is not discriminative enough. It assigns low values to models even if they have bad final ``loss''. As a result, minimizing that loss is not sufficient to guarantee a good solution. The third task loss (\textcolor{blue}{blue}) is a good task loss because it ``tracks'' the final goal ``loss'' well. In particular, it provides a \textit{strong learning signal} towards the range of acceptable models and is just \textit{discriminative} enough with respect to the final goal.

\subsection{Parallels between Predictive and Generative Tasks.\label{sec:ParallelsGenerative}}

Below, we contrast approaches in predictive (classification and structured prediction) and generative tasks (sampling and likelihood evaluation) with respect to their use of \textit{harder} and \textit{softer} task losses.

\paragraph{Predictive Tasks.} We contrast multi-class classification with structured prediction. Assume an unknown distribution $p(\x,\y)$. The goal is to learn a function $h_\theta:\X\to\Y$ which outputs a ``correct'' prediction $\y$ given an input $\x$. In multi-class classification, $\Y$ is a finite label space, and all mistakes are penalized equally according to the classification error, also known as the 0-1 loss:
\begin{align}
L_p(\theta)  = \E_{\x,\y\sim p(\x,\y)} \left[ 1_{\lbrace h_\theta(x_i) \neq y_i \rbrace} \right]
\end{align}
The 0-1 loss is a \textit{hard} loss because the only correct prediction is the ground truth.
In contrast, in structured prediction, $\Y$ is a structured space such as sentences, segmentation maps or bounding boxes, and can exponentially large or even infinite.
Mistakes are penalized according to a \textit{structured loss} $\ell(y',y;x)$:\footnote{Depending on the context, both $L_p(a)$ and $\ell$ are called task losses, as they implicitly define the task.}
\begin{equation}\label{eq:structuredpred}
	L_p(\theta) = \E_{(\x,\y) \sim p} \left[ \ell(h_\theta(\x),\y,\x) \right]
\end{equation}
Although the 0-1 loss $\ell(y',y;x)=1_{\lbrace h_\theta(x_i) \neq y_i \rbrace}$ could also be used, it is not informative of the type of mistake made, and can be detrimental to learning (Section~\ref{sec:hardsoftstructuredpred}). Instead, people often resort to \textit{softer} losses, such as the Hamming loss, which penalize mistakes more gradually, and result in better learning guarantees (Section~\ref{sec:hardsoftstructuredpred}).

\paragraph{Generative Tasks.} We contrast traditional likelihood-based generative modeling with GAN-style generative modeling based on parametric divergences. Assume an unknown target distribution $p(\x)$.
We focus on the (ill-defined) final task of learning a model $q_\theta(\x)$ which can generate ``realistic'' samples from ``the same'' distribution as $p$.
Traditional generative modeling formalize this task by penalizing models according to the negative log-likelihood:
\begin{equation}\label{eq:maxlikelihood}
	L_p(\theta)  = \E_{\x\sim p} \left[ - \log (q_\theta(\x)) \right]
\end{equation}
Although log-likelihood has been the \textit{de facto} learning objective in the past, 
there is no guarantee that log-likelihood is a good proxy for sample quality. In fact, \citet{theis2015note} have exhibited image models which have high likelihood but produce low quality samples, as well as models which have low likelihood but produce high quality samples. 
Maximizing the log-likelihood is equivalent to minimizing the Kullback-Leibler divergence 
\begin{align} \mathbf{KL}(p||q_\theta)=\E_{\x\sim p} \left[ \log (\frac{p(\x)}{q_\theta(\x)}) \right]\end{align}
which can be considered a \textit{hard} loss, in the sense that mistakes are penalized regardless of the metric structure of $\X$. This is particularly obvious in the special case where $p(x)=\delta(x-x_p)$ and $q(x)=\delta(x-x_q)$ are two Dirac distributions. Then, $KL(p||q_\theta)$ equals 0 if $x_p=x_q$ and infinity otherwise, which means all mistakes are penalized equally regardless of the distance between $x_p$ and $x_q$. There are ways to make the maximum likelihood ``softer'', but they come with their own caveats (Section~\ref{sec:interactionskl}).

GAN-style models adopt a very different approach and penalize models according to a parametric divergence :
\begin{align}
L_p(\theta)  =& \Div(p||q_\theta) = \sup_{f\in\F} \E_{(\x,\x')\sim p\otimes {q_\theta}}[\Delta(f(\x), f(\x'))]
\end{align}
For instance, the saturating GAN can be formulated as the minimization of a parametric Jensen-Shannon divergence:
\begin{align}
L_p(\theta)  =& \Div_{\text{ParamJS}}(p||q_\theta) = \sup_{f\in\text{NeuralNet}} \E_{\x\sim p}[\log f(\x)] + \E_{\x\sim q}[\log (1-f(\x))] \, .
\end{align}
In general, the parametric divergence only penalizes moments which are captured by the discriminator class.
This means that we can arbitrarily tune the ``hardness'' of the parametric divergence by tuning $\F$ and $\Delta$.
As an extreme example, if $\F$ is the set of linear 1-Lipschitz functions and we consider the Wasserstein-1 formulation $\Delta(f(x),f(x')) = f(x)-f(x')$, then the resulting parametric divergence reduces to the distance between the distribution means $||\E_{\x\sim p}[x] - \E_{\x\sim q}[x]||$. This is substantially \textit{softer} than the corresponding (nonparametric) Wasserstein obtained by removing the linear constraint.
In general, parametric divergences are promising task losses for generative modeling as long as we can find ways to tailor the discriminator to approximate the notion of realistic samples well.

\subsection{An Analogy with Structured Prediction\label{sec:hardsoftstructuredpred}}

We derive an intuitive analogy between ``hard'' losses such as the 0-1 loss and the KL-divergence and ``softer'' losses like the Hamming loss and the Wasserstein distance.
Then, we draw insights from the convergence results of \citet{osokin2017structured} in structured prediction, which parallel the intuition in generative modeling that learning with weaker\footnote{In the topological sense.} divergences is easier~\citep{arjovsky2017wasserstein} and more intuitive~\citep{liu2017approximation} than with stronger divergences.

\paragraph{Analogy between Structured Prediction Losses and Divergences.}
A loose analogy can be made between ``hard'' losses like the 0-1 loss and the KL-divergence.
A similar analogy can be made between ``softer'' losses like the Hamming loss and the Wasserstein distance.
Consider two Dirac distributions $p(y)=\delta(y-y_p)$ and $q(y)=\delta(y-y_q)$ defined over~$\R^D$.
We compute the Jensen-Shannon divergence, which can be thought of as a symmetrized KL:
$$\mathbf{JS}(p||q)=\frac 1 2 \mathbf{KL}(p||\frac{p+q}{2}) + \frac 1 2 \mathbf{KL}(q||\frac{p+q}{2}) = 1_{\lbrace y_p\neq y_q \rbrace} \cdot \log 2 \, .$$
In this case, the Jensen-Shannon divergence reduces to a scaled 0-1 loss between the atoms.
We now consider the Wasserstein distance $\mathbf{W}(p,q)$ with base distance $L_2$, which yields here:
$$\mathbf{W}(p,q)=||y_p-y_q||_2 \, .$$
The Wasserstein distance reduces to the $L_2$ distance between the atoms, or equivalently the Hamming distance, if we consider only binary vectors.
In that sense, the KL-divergence could be considered a ``hard'' divergence, while the Wasserstein distance could be considered a ``softer'' divergence.

\paragraph{``Softer'' losses are better in structured prediction.} Consider a ``hard'' structured loss, the 0-1 loss, defined as $\ell_{0-1}(\y,\y') \widehat= \bm{1} \left\{\y \neq \y'\right\}$, and a ``softer'' loss, the Hamming loss, defined as $\ell_\text{Ham}(\y,\y') \widehat= \frac{1}{T}\sum_{t=1}^T \bm{1}\{\y_t \neq \y'_t\}$, when $\y$ decomposes as $T=\log_2 |\Y|$ binary variables $(y_t)_{1\leq t\leq T}$. Because ``softer'' losses like the Hamming loss are more informative about the mistakes, we can expect that fewer examples are needed to learn the model.
This is, with caveats, what was shown by~\citet{osokin2017structured} in a nonparametric setting.\footnote{The analysis of \citet{osokin2017structured} is nonparametric in the sense that it ignores the dependence on $x$ (it allows an arbitrary dependence on $x$ for the score functions by using an infinite dimensional RKHS function space). Additionally, they only consider convex consistent surrogate losses in their analysis, and they give upper bounds but not lower bounds on the sample complexity. It is possible that optimizing approximately-consistent surrogate losses instead of consistent ones, or making additional assumptions on the distribution of the data could yield better sample complexities.} \citet{osokin2017structured} derive a worst case sample complexity needed for a simple learner based on surrogate loss minimization via stochastic gradient descent to achieve a fixed regret $\epsilon>0$ with respect to the best generalization error possible. The sample complexity quantifies how many samples are needed for the simple learner to guarantee a regret of $\epsilon$. When the task is the 0-1 loss, they get a sample complexity of $O(|\Y| / \epsilon^2)$, which is exponential in the dimension of $y$. However, when the task loss is the Hamming loss, they get a much better sample complexity of $O(\log_2 |\Y| / \epsilon^2)$ which is linear in the number of dimensions, whenever certain constraints are imposed on the score function~\citep[see][section on exact calibration functions]{osokin2017structured}. In contrast, if the surrogate loss of the learner is based on the 0-1 loss (the analog of the top red curve in Figure~\ref{fig:goodandbad}), but the task loss is the Hamming loss (giving rise to the analog of the bottom green curve in Figure~\ref{fig:goodandbad}), than the sample complexity of the learner is still exponential.

Thus their results suggest that choosing the \textit{right} structured loss, like the ``softer'' Hamming loss, might make training \textit{exponentially} faster. According to the previous analogy, these results could mean that it might also be more efficient to use softer losses than the KL-divergence for training generative models.
These observations echo results in generative modeling \citep{arjovsky2017wasserstein, liu2017approximation} showing that it can be easier to learn with weaker divergences than with stronger ones (in the topological sense). In particular, one of their arguments is that distributions with disjoint support can be compared in weaker topologies like the one induced by the Wasserstein but not in stronger ones like the one induced by the Jensen-Shannon. 

However, we will show in Section~\ref{sec:samplecomplexity} that the Wasserstein distance has other issues, such as poor sample complexity. 
On the other hand, parametric adversarial divergences, which are also a softer alternative to the KL-divergence, have reasonable sample complexity and other good properties which are discussed in Sections~\ref{sec:scale}, \ref{sec:reflecttask}~and~\ref{sec:pmi}.

\subsection{Training and Evaluation in Practice\label{sec:practice}}

%

%
%
%
%
%
%
%
%

%
%
\paragraph{Training vs. Evaluation Loss.}

Strictly speaking, task losses should be regarded as evaluation metrics,
from which we can then derive training losses which are easier to optimize (e.g. 0-1 task loss is approximated with cross-entropy training loss).
In practice, parametric divergences are only used as training losses but not as evaluation metrics, because optimization instability makes it difficult to compute them reliably. However, future work might make it possible to use parametric divergences as evaluation metrics as well.

\paragraph{Estimation Biases.}

Consider two fixed distributions $p$ and $q$. We are usually interested in the \textit{population} divergence
\begin{align}
L_p(\theta) \widehat =& \Div(p||q_\theta) = \sup_{f\in\F} \E_{(\x,\x')\sim p\otimes {q_\theta}}[\Delta(f(\x), f(\x'))]
\end{align}
In practice, $p$ is unknown and we only have access to samples. We might also only have access to samples from the model when it is an implicit one. 
Denoting $D_{train} = \lbrace \x_{train}^{(i)} \overset{\text{iid}}{\sim} p, \; \y_{train}^{(i)} \overset{\text{iid}}{\sim} q\rbrace$, we compute the \textit{training} divergence as follows :
\begin{align}\label{eq:trainingdivergence}
\Div(p||q_\theta)_{train} = \sup_{f\in\F} \frac{1}{N} \sum_{i=1}^N \Delta(f(\x_{train}^{(i)}), f(\y_{train}^{(i)}))
\end{align}
Denote $\widehat f_{ERM}$ the previous minimizer. We define the \textit{validation} divergence as the evaluation of $\widehat f_{ERM}$ over a validation set $D_{val} = \lbrace \x_{val}^{(i)} \overset{\text{iid}}{\sim} p, \;  \y_{val}^{(i)} \overset{\text{iid}}{\sim} q\rbrace$:
\begin{align}
\Div(p||q_\theta)_{val} = \frac{1}{N'} \sum_{i=1}^{N'}   \Delta(\widehat f_{ERM}(\x_{val}^{(i)}), \widehat f_{ERM}(\y_{val}^{(i)}))
\end{align}
Analogously to usual classification bounds (the discriminator can be seen as a classifier), the expected training divergence is \textit{higher} than the population divergence, while the expected validation divergence is \textit{lower} than the population divergence, where the expectations are taken over the sampling of $D_{train},D_{val}$.

\begin{theoremtwo}[Parametric Divergence Biases] The following inequalities hold :
\begin{align}\label{eq:bounds}
\underbrace{\E_{D_{train},D_{val}} [\Div(p||q)_{val}]}_{\text{validation divergence}} \leq \underbrace{\Div(p||q)}_{\text{population divergence}} \leq \underbrace{\E_{D_{train}}[\Div(p||q)_{train}]}_{\text{training divergence}}
\end{align}
\end{theoremtwo}

\begin{proof}
The second inequality results from the fact that taking the supremum inside the expectation $\E_{D_{train}}[\cdot]$ is always higher than outside it.
\begin{align}
 \sup_{f\in\F} \underbrace{ \E_{D_{train}}[ \frac{1}{N} \sum_{i=1}^N \Delta(f(\x_{train}^{(i)}), f(\y_{train}^{(i)}))]}_{ \E_{(\x,\x')\sim p\otimes {q_\theta}}[\Delta(f(\x), f(\x'))]} \leq  \E_{D_{train}}[\underbrace{ \sup_{f\in\F} \frac{1}{N} \sum_{i=1}^N \Delta(f(\x_{train}^{(i)}), f(\y_{train}^{(i)}))}_{\Div(p||q)_{train}} ]
\end{align}
For the first inequality, taking the expectation of the validation divergence with respect to the sampling of the validation set gives
\begin{align}
\E_{D_{val}} [\frac{1}{N} \sum_{i=1}^N   \Delta(\widehat f_{ERM}(\x_{val}^{(i)}), \widehat f_{ERM}(\y_{val}^{(i)}))] =& \E_{(\x,\x')\sim p\otimes {q_\theta}}[\Delta(\widehat f_{ERM}(\x), \widehat f_{ERM}(\x'))] \\&\leq  \sup_{f\in\F} \E_{(\x,\x')\sim p\otimes {q_\theta}}[\Delta(f(\x), f(\x'))]
\end{align}
where the inequality comes from the definition of the supremum. Now, we take expectations with respect to the sampling of the training set, which $\widehat f_{ERM}$ depends on:
\begin{align}
\E_{D_{train},D_{val}} [\underbrace{ \frac{1}{N} \sum_{i=1}^N   \Delta(\widehat f_{ERM}(\x_{val}^{(i)}), \widehat f_{ERM}(\y_{val}^{(i)})) }_{\Div(p||q)_{val}}] \leq \underbrace{ \sup_{f\in\F} \E_{(\x,\x')\sim p\otimes {q_\theta}}[\Delta(f(\x), f(\x'))] }_{\Div(p||q)}
\end{align}
\end{proof}

Notice that the validation divergence is not an unbiased estimator of the population divergence, because the discriminator was only optimized over the training set.
Therefore, it is always worthwhile to look both at training and validation divergence to bound the population divergence. If the discriminator is overfitting to the training set, then there would be a large \textit{generalization} gap between the two, while if the values are close, then we have a good estimate of the population divergence.
To the best of our knowledge, there is no unbiased estimators of the population divergence in the general case. An exception is the family of f-divergences which can be estimated with Monte-Carlo in the special case where $p,q$ have a density which can be evaluated. Another exception is MMD~\citep{bellemare2017cramer}, which admits a closed-form unbiased estimator. However, such estimator is only unbiased for a fixed kernel. If the kernel were to be optimized to maximize some discrepancy between empirical distributions, in the same way a discriminator is optimized, then the MMD estimator would become biased as well~\citep{binkowski2018demystifying}.

\section{Good Divergences Should Scale Well with Data Dimensionality \label{sec:scale}}

One of the main difficulties in generative modeling is to deal with high-dimensional data because of the curse of dimensionality, which states that in order to fill a given volume in data space, it takes an amount of data which is exponential in its dimension.
The sample complexity, which characterizes how much data is needed to approximate a given divergence, is discussed for nonparametric and parametric divergences in Section~\ref{sec:samplecomplexity}. We also compare some nonparametric and parametric divergences experimentally in Sections~\ref{sec:sinkhornexperiment}~and~\ref{sec:thin8}.

\subsection{Theoretical Sample Complexities in High-Dimensions\label{sec:samplecomplexity}}

Consider two distributions $p,q$ and their associated empirical distributions $\widehat p_n, \widehat q_n$. Typically, $p$ is the unknown distribution to learn, and $q$ is the model (or generator) distribution.
How much data is required to approximate the true (population) divergence $\Div(p||q)$ with the empirical divergence $\Div(\widehat p_n||\widehat q_n)$? Formally, we can define the sample complexity as the minimal number of samples $n$ such that $\lvert \Div(p||q) - \Div(\widehat p_n||\widehat q_n) \rvert \leq \epsilon$ with high probability for $\epsilon>0$. 

Following the terminology of~\citet{mohamed2016learning}, we distinguish the case of \textit{explicit} model, where the density $q(x)$ can be numerically evaluated, and the case of \textit{implicit} model, where it is only possible to sample from $q$. For instance, GAN generators are implicit models, and have a distribution which is typically supported in a low-dimensional manifold and does not admit a density with respect to the usual measure. Explicit models are more restrictive and need to have a full-dimensional support. For instance, VAEs and PixelCNNs are explicit models, but they each come with their own problems (see Section~\ref{sec:interactionskl}). We summarize the sample complexities for some parametric and nonparametric divergences in \textbf{Table~\ref{fig:convcomp}}.

\begin{table*}[h] %
\begin{center}
\centerline{
\resizebox{1.\textwidth}{!}{
\begin{tabular}{|llll| }
 \hline
 Divergence & Sample Complexity & Computation & Tunable to Final Task\\
 \hline
 \multicolumn{4}{|c|}{Explicit model: Can evaluate $p(x)$}\\
 \hline
 f-Divergences & $O(1/\epsilon^2)$ & Monte-Carlo, $O(n)$ &  No\\
  \hline
 \multicolumn{4}{|c|}{Implicit model: Can only sample $x\sim p$ }\\
 \hline
 f-Divergences & \multicolumn{3}{c|}{undefined in general} \\
 Nonparametric Wasserstein & $O(1/\epsilon^{d/2})$ & Sinkhorn, $O(n^2)$ &   in base distance \\  $\text{MMD}$ & $O(1/\epsilon^2)$ & analytic, $O(n^2)$& in kernel \\
 Parametric Divergence & $O(p/\epsilon^2)$ & SGD & in discriminator\\
  Parametric Wasserstein Div. & $O(p/\epsilon^2)$ & SGD & in discriminator \& base distance\\
 \hline
\end{tabular}
} %
} %
\caption{\small 
Properties of Divergences. Note that although f-divergences can be estimated efficiently for explicit models, they are usually not defined for implicit models (see text). MMD can be estimated efficiently in closed form and can be tuned through the choice of kernel, but is known to lack discriminative power for generic kernels. The nonparametric Wasserstein can be computed iteratively with the Sinkhorn algorithm, and can integrate the final loss in its base distance, but requires exponentially many samples to estimate which makes it impractical in high dimensions. Parametric divergences have reasonable sample complexities, can be computed iteratively with SGD, and can integrate the final loss through the choice of class of discriminators and the choice of side-tasks. In particular, the parametric Wasserstein has the additional possibility of integrating the final loss into the base distance.
\label{fig:convcomp}
}
\end{center}
\vspace*{-4mm}
\end{table*}

\paragraph{Parametric Divergences.} Parametric adversarial divergences can be formulated as a classification problem between $p$ and $q$, with a loss depending on the specific adversarial divergence. They can be estimated for implicit models and have a reasonable sample complexity of $O(p/\epsilon^2)$, where $p$ is the VC-dimension/number of parameters of the discriminator~\citep{arora2017generalization}. They are usually computed using stochastic gradient descent (SGD), which provides no guarantees of finding a global minimum. This is not necessarily a bad thing, as it has been theoretically shown in the supervised case that SGD induces some form of bias or regularization, leading for instance to maximum-margin solutions in the separable-data case~\citep{soudry2018implicit}. We might expect the same type of implicit regularization to be beneficial for parametric divergences too.

\paragraph{(Nonparametric) Wasserstein.} A straightforward estimator of the (nonparametric) Wasserstein is simply the Wasserstein distance between the empirical distributions $\widehat p_n$ and $\widehat q_n$, for which smoothed versions can be computed in $O(n^2)$ using specialized algorithms such as Sinkhorn's algorithm~\citep{cuturi2013sinkhorn} --it is then known as the plug-in estimator-- or iterative Bregman projections~\citep{benamou2015iterative}. However, the empirical Wasserstein is a biased estimator and has sample complexity $n=O( 1 / \epsilon^{d})$ which is exponential in the number of dimensions~\citep[see][Corollary 3.5]{sriperumbudur2012empirical}. This bound was recently improved to $O(1/\epsilon^{d/2})$ by \citet{chizat2020faster}. Thus, the theory suggests that empirical Wasserstein is not a viable estimator in high-dimensions. In practice, we compare nonparametric and parametric Wasserstein on two generation tasks in Section~\ref{sec:sinkhornexperiment}.

\paragraph{f-divergences.} For explicit models which allow evaluating the density $q_\theta(x)$, one could use Monte-Carlo to evaluate the f-divergence with sample complexity $n=O(1/\epsilon^2)$, according to the Central-Limit theorem.  However, for implicit models, there is no one good way of estimating f-divergences from samples. And in fact, most f-divergences are generally not defined for empirical distributions because they might not be absolutely continuous with another. There are some techniques for estimation~\citep{nguyen2010estimating,moon2014multivariate,ruderman2012tighter}, but they all make additional assumptions about the underlying densities (such as smoothness), or they solve the dual in a restricted family, such as a RKHS, which makes the divergences no longer f-divergences.

\paragraph{MMD.} Maximum Mean Discrepancy admits an estimator with sample complexity $n=O(1/\epsilon^2)$, which can be computed analytically using U-statistics $O(n^2)$, or even in $O(n)$ using the linear estimator~\citet{gretton2007kernel}. One should note that MMD depends fundamentally on the choice of kernel. As the sample complexity is independent of the dimension of the data, one might believe that the MMD estimator behaves well in high dimensions. However, it was experimentally illustrated in~\citet{dziugaite2015training} that with generic kernels like RBF, MMD performs poorly for MNIST and Toronto face datasets, as the generated images have many artifacts and are clearly distinguishable from the training dataset.

It was shown theoretically in \citep{reddi2014decreasing} that the power of the MMD statistical test can drop polynomially with increasing dimension, which means that with generic kernels, MMD might be unable to discriminate well between high-dimensional generated and training distributions.
Specifically, consider a Gaussian kernel with bandwidth $\gamma$ and compute the $\text{MMD}^2$ between two isotropic Gaussians with different means. Then, for $ 0 < \epsilon \le 1/2$, and $d\longrightarrow\infty$, $\text{MMD}^2$ goes to zero: 
\begin{itemize}
\item polynomially as $1/d$ if $\gamma = \sqrt{d}$
\item polynomially as $1/d^{1+2\epsilon}$ if $\gamma = d^{1/2-\epsilon}$
\item exponentially as $\exp(d^{2\epsilon}/2)$ if $\gamma = d^{1/2+\epsilon}$, all that while the KL divergence between the two Gaussians stays constant.
\end{itemize}
which suggest MMD with fixed kernel cannot separate high-dimensional distributions well.

\paragraph{Limitations of Sample Complexity Analysis.}
Note that comparing divergences in terms of sample complexity can give good insights on what is a good divergence, but should be taken with a grain of salt as well. On the one hand, the sample complexities we give are upper-bounds, which means the estimators could potentially converge faster. On the other hand, one might not need a very good estimator of the divergence in order to learn in some cases. This is illustrated in our experiments with the nonparametric Wasserstein  which has bad sample complexity but yields reasonable results in some cases (Section~\ref{sec:sinkhornexperiment}).

\subsection{Nonparametric vs. Parametric Wasserstein (Experiment)\label{sec:sinkhornexperiment}}

\begin{figure}
\centering
\includegraphics[width=0.8\linewidth]{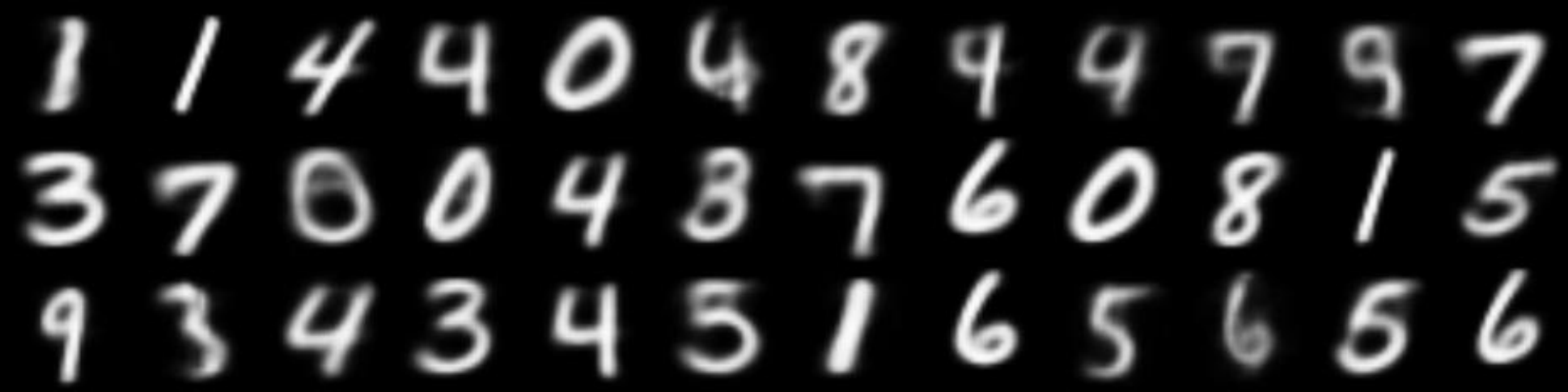}\\
\vspace{1em}
\includegraphics[width=0.8\linewidth]{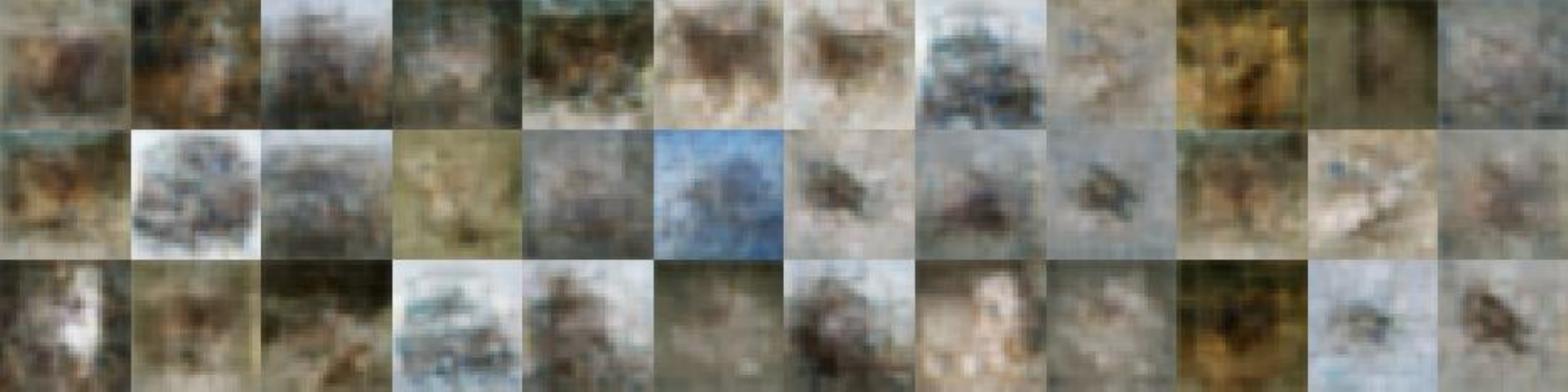}
\caption{\label{fig:sinkhron} \small Images generated by the network after training with the Sinkorn-Autodiff algorithm on MNIST dataset (top) and CIFAR-10 dataset (bottom). One can observe than although the network succeeds in learning MNIST, it is unable to produce convincing and diverse samples on the more complex CIFAR-10.}
\end{figure}

Since the sample complexity of the nonparametric Wasserstein  is exponential in the dimension (Section~\ref{sec:samplecomplexity}), we verify experimentally whether training a generator to minimize the \emph{nonparametric Wasserstein} distance works in high dimensions. We implement the Sinkhorn-AutoDiff algorithm~\citep{genevay2017sinkhorn} to compute the entropy-regularized $L_2$-Wasserstein distance between minibatches of training images and generated images, and minimize the divergence using stochastic gradient descent.\footnote{\url{https://github.com/gpeyre/SinkhornAutoDiff}} \textbf{Figure \ref{fig:sinkhron}} shows generated samples after training with the Sinkhorn-Autodiff algorithm on both MNIST and CIFAR-10 dataset. We then minimize the estimated divergence using stochastic gradient descent. On MNIST, the network manages to produce decent but blurry images. However on CIFAR-10, which has higher intrinsic dimensionality, the generator fails to produce meaningful samples. This is in stark contrast with the high quality generators displayed in the literature with a \emph{parametric  Wasserstein} (Wasserstein-GAN). This result would suggest that indeed the nonparametric Wasserstein should not be used for generative modeling when the (effective) dimensionality is high.

\subsection{Generating Simple High-dimensional Images (Experiment) \label{sec:thin8}}

\begin{figure}

\centering

\begin{minipage}{0.07\linewidth}
\scriptsize \textbf{Train}
\end{minipage}
\begin{minipage}{0.92\linewidth}
\includegraphics[width=0.32\linewidth]{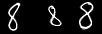}
\includegraphics[width=0.32\linewidth]{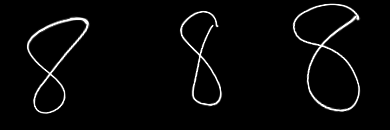}
\includegraphics[width=0.32\linewidth]{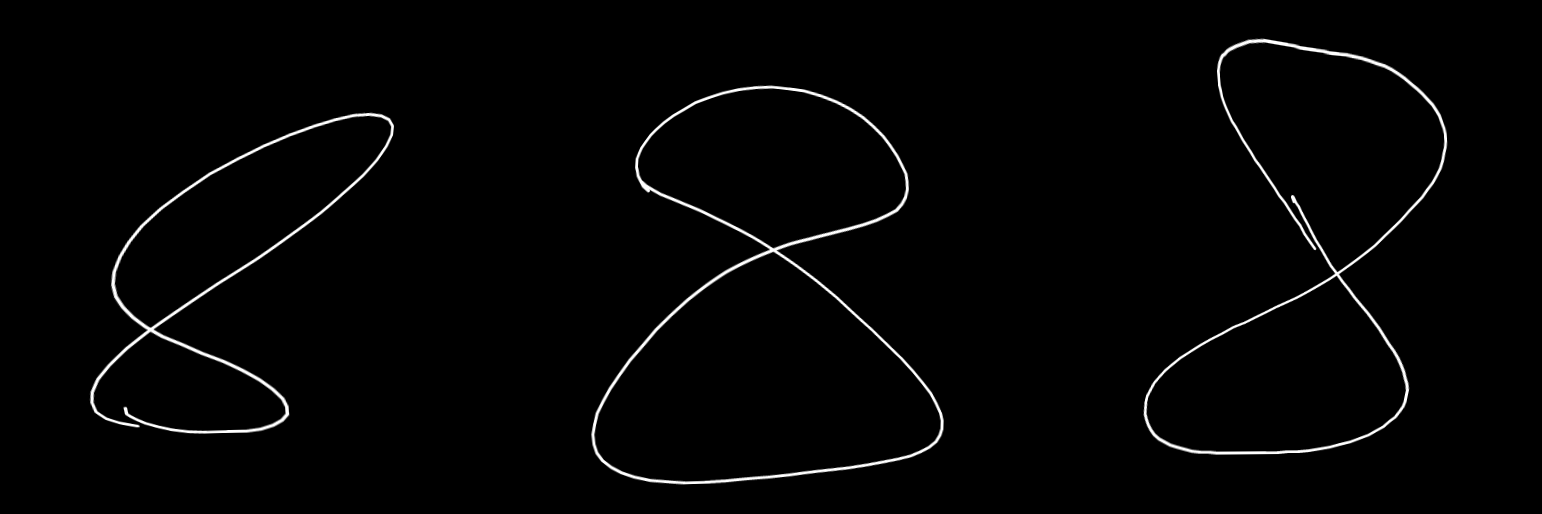}
\end{minipage}

\begin{minipage}{0.07\linewidth}
\scriptsize GAN
\end{minipage}
\begin{minipage}{0.92\linewidth}
\includegraphics[width=0.32\linewidth]{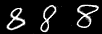}
\includegraphics[width=0.32\linewidth]{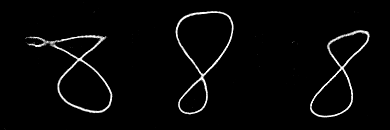}
\includegraphics[width=0.32\linewidth]{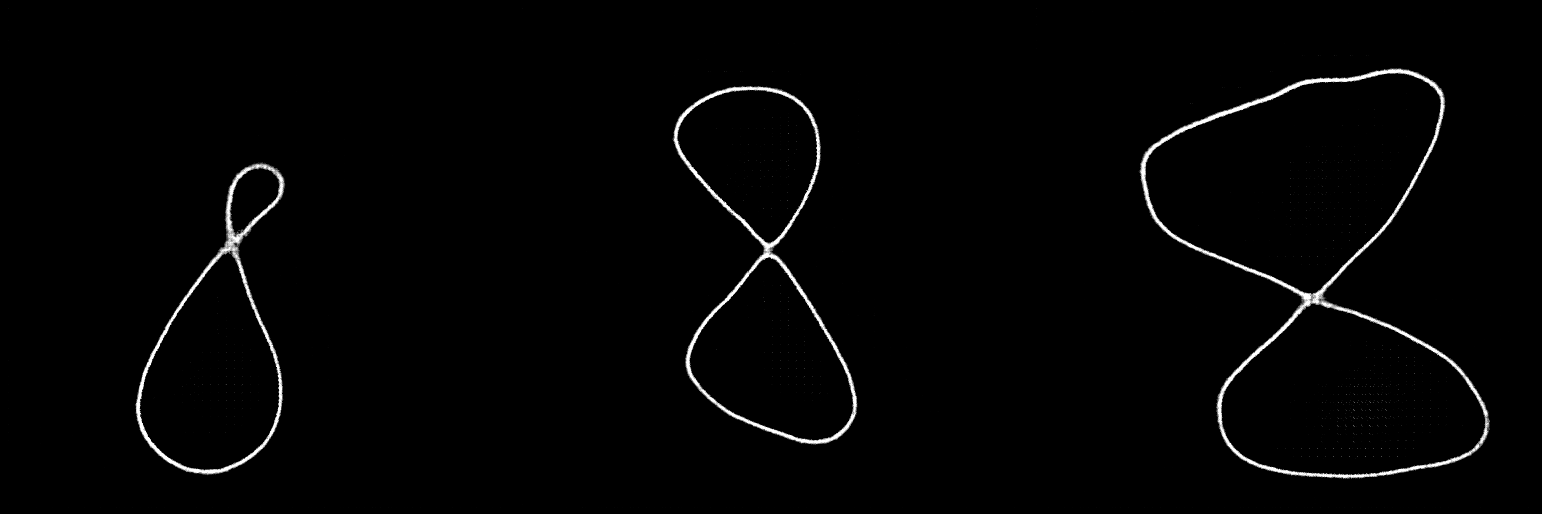}
\end{minipage}

\begin{minipage}{0.07\linewidth}
\scriptsize VAE
\end{minipage}
\begin{minipage}{0.92\linewidth}
\includegraphics[width=0.32\linewidth]{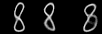}
\includegraphics[width=0.32\linewidth]{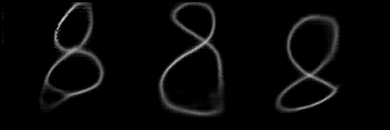}
\includegraphics[width=0.32\linewidth]{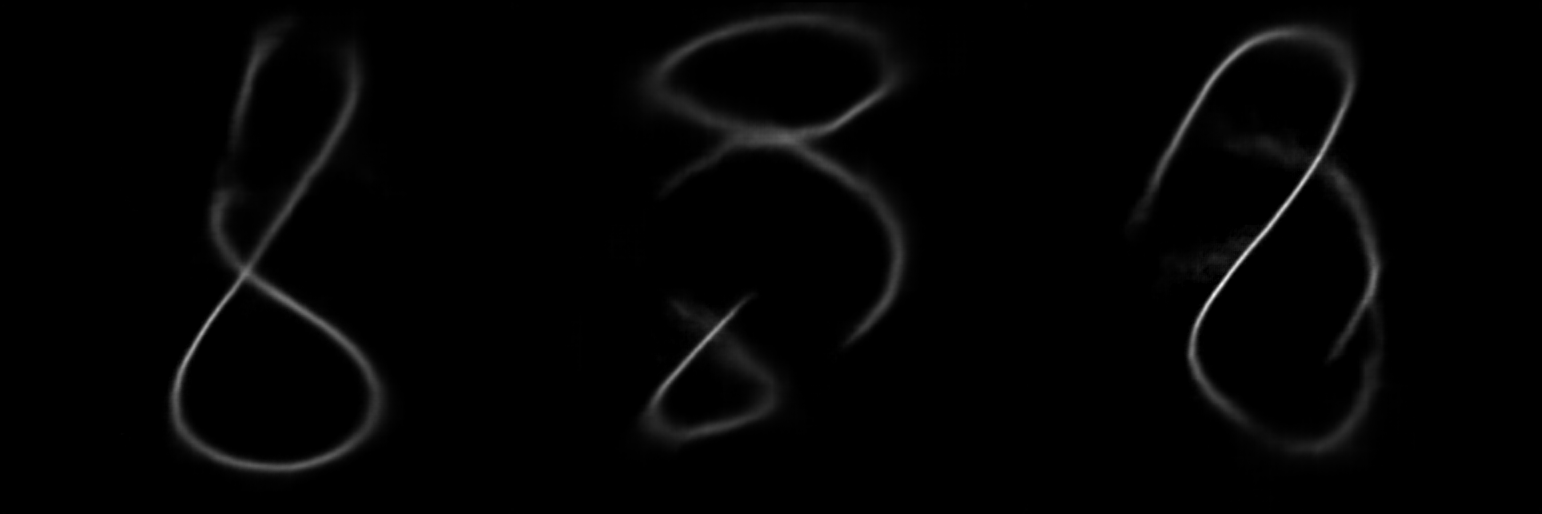}
\end{minipage}

\caption{\small Samples from \emph{Thin-8} training set (\textbf{top row}), WGAN-GP (\textbf{middle row}) and Convolutional VAE (\textbf{bottom row}) with 16 latent variables. Resolutions are $32\times 32$ (\textbf{left column}), $128\times 128$ (\textbf{middle column}), and $512\times 512$ (\textbf{right column}). Note how the GAN samples are always crisp and realistic across all resolutions, while the VAE samples tend to be blurry with gray pixel values in high-resolution. We can also observe some averaging artifacts in the top-right 512x512 VAE sample, which looks like the average of two ``8''. More samples can be found in Section~\ref{sec:mil8-more} of the Appendix.\label{fig:mila8}}
\end{figure}

There has recently been a number of successes in modelling high-dimensional images with GANs, such as $1024\times 1024$ faces~\citep{karras2017progressive} and $512\times 512$ photos~\citep{brock2018large}, which does suggest that parametric divergences are very successful in modelling high-dimensional data. We collect \emph{Thin-8}, a dataset of 1585 grayscale handwritten images of the digit 8, with a very high resolution of $512\times 512$.\footnote{Thin-8 dataset can be download from \url{https://gabrielhuang.github.io/\#thin}} The \emph{Thin-8} task differs from the aforementioned tasks because the images to model have \textbf{very low intrinsic dimensionality}  (each digit is one curve which can be parametrized using only a handful of points), which allows us to factor out the usual complexity of high-dimensional photos. Therefore, even a simple convolutional network with few filters and few latent-dimensions (16 in our experiments) should be able to generate the images.

We train\footnote{Code available at \url{https://github.com/gabrielhuang/adversarial-divergence-code}} a convolutional VAE and a WGAN-GP~\citep{gulrajani2017improved}, henceforth simply denoted GAN, using nearly the same architectures (VAE decoder similar to GAN generator, VAE encoder similar to GAN discriminator), with 16 latent variables, on the following resolutions: $32\times 32$, $128\times 128$ and $512\times 512$. We optimize the losses using Adam, and augment the samples with random elastic deformation during training.

Generated samples are shown in \textbf{Figure~\ref{fig:mila8}}. We observe that the VAE, trained to minimize the evidence lower bound on maximum-likelihood, fails to generate convincing samples in high-dimensions: they are blurry, pixel values are gray instead of being white, and some samples look like the average of many digits. This can be explained by the fact that the smoothing used in the VAE reduces the maximum likelihood to a pixel-wise reconstruction loss, which is problematic here. Indeed, because the images are dominated by background pixels and the strokes are very thin, with high probability, any two ``8' will intersect on no more than a little area, so pixel-wise distances are not meaningful.

On the contrary, the GAN which is trained with a parametric Wasserstein distance can generate sharp and realistic samples even in $512\times 512$. Our hypothesis is that the discriminator learns only the moments that matter for visual quality, and that matching only those moments is easier than matching all of the moments, which is required for really maximizing the likelihood. 

In conclusion, the high-dimensional aspect was clearly an obstacle for successfully training a fairly simple generator using maximum-likelihood, while using a parametric divergence to train the \textit{exact} same generator allowed to overcome the high-dimensional aspect and resulted in high-quality samples.

\section{Good Divergences Should Reflect the Final Task\label{sec:reflecttask}}
 
In Section~\ref{sec:priortask}, we discussed the necessity and importance of designing task losses which reflect the final task. We showed that in structured prediction, optimizing more informative task losses can make learning considerably easier under some conditions. Similarly, in generative modeling, we would like divergences to be as informative and close to the final task as possible. Although not all divergences can easily integrate final task-related criteria, parametric divergences can be tuned through side-tasks and indirectly through their architecture (Section~\ref{sec:flex}).  We study the synthetic task of generating images of digits that sum up to 25, and compare KL-based and parametric divergences in their ability to enforce that constraint (Section~\ref{sec:25controlled}~and~\ref{sec:25e2e}).

\subsection{Tuning Various Divergences to the Final Task\label{sec:flex}}

Pure f-divergences cannot directly integrate \textit{any} notion of final task. By default, f-divergences might not have the expected properties, as we show experimentally in Section~\ref{sec:25controlled}. To some extent, there is a possibility of tweaking f-divergences by combining them with generators that have a \textit{special structure}; this is discussed in Section~\ref{sec:interactionskl}. One could also attempt to induce properties of interest by adding a \textit{regularization} term to the f-divergence. However, if we assume that maximum likelihood is itself often not a meaningful task loss, then there is no guarantee that minimizing a tradeoff between maximum likelihood and a regularization term is more meaningful or easier.

The Wasserstein distance and MMD are respectively induced by a base metric $d(\x,\x')$ and a kernel $K(\x,\x')$. The metric and kernel give us the opportunity to specify a task by letting us express a (subjective) notion of \textit{similarity}. However, the metric and kernel traditionally had to be defined by hand. For instance, \citet{genevay2017sinkhorn} learn to generate MNIST by minimizing a smooth Wasserstein based on the $L_2$-distance, while \citet{dziugaite2015training, li2015generative} also learn to generate MNIST by minimizing the MMD induced by kernels obtained externally: either generic kernels based on the $L_2$-distance or on autoencoder features. However, the results seems to be limited to simple datasets. There is no \emph{obvious} or generally accepted way to learn the metric or kernel in an end-to-end fashion; this is an active research direction.
In particular, MMD has recently been combined with aversarial kernel learning, with convincing results on LSUN, CelebA and ImageNet images: \citet{mroueh2017mcgan} learn a feature map and try to match its mean and covariance, \citet{li2017mmd} learn kernels end-to-end, while \citet{bellemare2017cramer} do end-to-end learning of energy distances, which are closely related to MMD. See~\citet{binkowski2018demystifying} for a recent review of MMD-based GANs.

\emph{Parametric} adversarial divergences can be tweaked to fit the final task in several ways. 
The first knob is the choice of discriminator \textit{architecture}, which implicitly determines what aspects of the data the divergence is more sensitive or blind to. A typical choice of discriminator architecture for image generation are  convolutional neural networks~\citep{radford2015unsupervised}, since CNNs have several good properties for assessing whether an image is realistic: ability to detect blurriness, edges and textures, while being robust to translations and small deformations. 
The second knob is the use of a \textit{side-task}. Instead of solely training the discriminator to distinguish true and generated data, one can also train the discriminator to solve a relevant side-task at the same-time, with the hope that it induces properties of interest on the resulting divergence. We will show in Sections~\ref{sec:25controlled}~and~\ref{sec:25e2e} how a discriminator can be made much more sensitive to certain aspects of the data using a side-task.
 
\subsection{Sensitivities of Divergences to Different Aspects of the Sum-25 Distribution\label{sec:25controlled}}

In this section, we introduce the Sum-25 task as a benchmark for comparing the sensitivities of divergences to different aspects of the data. We then add a side-task to a parametric divergence and show that it can improve sensitivity to important aspects of the data. 

\paragraph{Sum-25 Task.} 

The Sum-25 task consists in generating combinations of 5 digits that sum up to 25. We devise the following, \textit{on-the-fly} dataset. First, we enumerate all 5631 combinations of 5 digits (out of 100,000) such that these digits sum up to 25. Then, we split them into disjoint train (50\%) and test (50\%) sets. The sampling process consists in uniformly sampling a random combination from the train/test set, then sampling corresponding digit images from MNIST, and finally concatenating them to yield the final image containing the 5 digits in a row summing up to 25. 

\paragraph{Visual and Symbolic Constraints.}  
The Sum-25 task can be thought of as a toy version of more complex tasks which have their own constraints. An example real task could be to generate pictures which satisfy perspective (e.g. objects must be skewed appropriately) and physical constraints (e.g., no floating objects). In the case of the Sum-25 task, the generator needs to model \textit{two} aspects of the data accurately:
\begin{itemize}
\item \textit{Visual constraint} : Digits should be recognizable and have good visual quality.
\item \textit{Symbolic constraint }: Digits must sum up to 25. 
\end{itemize}
Therefore, can we define divergences which enforce both visual and symbolic constraints~?

\paragraph{Factorized Distributions.} %
To factor out the effects of the learning process, we compute divergences between fixed reference $p$ and candidate $q$ distributions. Additionally, we restrict $p$ and $q$ distributions to be factorizable into a symbolic model and a conditional visual model as follows
\begin{align}\label{eq:factored}
q(x) = \sum_z q(x,z) = \sum_z \underbrace{q(z)}_{\text{symbolic model}} \prod_{i=1}^5 \underbrace{q(x_i | z_i)}_{\text{conditional visual model}}
\end{align}
where $z_i$ are symbolic digits and $x_i$ images of digits. 
We denote $\Test$-25 the reference distribution, which is uniform over combinations of 5 digits from the MNIST test-set which sum up to 25 (symbolic constraint satisfied). 

\paragraph{Sensitivity of KL-divergence to Constraints.}

In our first experiment, we consider 6~candidate distributions based on combining 3 conditional single-digit visual models with 2~symbolic models. The visual models are VAEs which are trained with Adam to generate single MNIST digits for 10, 70, and 80 epochs (prefixes $\Vae_{10}$-, $\Vae_{70}$-, $\Vae_{80}$-).
The symbolic models \textbf{-25} (resp. \textbf{-Non25}) are the uniform distributions over combinations of 5 digits that sum up (resp. do no sum up) to 25.
We compute the $\E_p[-\log q]$ using~\eqref{eq:factored}, but replacing $q(x_i|z_i)$ with the variational lower-bound of the VAE, since these can be naturally computed. The numbers in\textbf{ Table~\ref{table:nllfactorized}} show that the negative log-likelihood is \textit{overly sensitive} to the visual appearance, but barely cares about the higher-level Sum-25 constraint. Therefore, by training with such an objective we might end up with digits that don't sum up to 25. More generally, this suggests that models trained with nonparametric divergences such as KL might not be able to enforce certain constraints which might be important to the final task.

\begin{table}
\centering
\begin{tabular}{|ccc|}
\hline
Visual Model $q(x_i|z_i)$ & Sum-25 satisfied & Sum-25 NOT satisfied \\
\hline
$\Vae_{10}$ & 572.3 $\pm$ 1.4 & 575.4 $\pm$ 1.5 \\
$\Vae_{70}$ & 488.9 $\pm$ 1.2 & 487.3 $\pm$ 1.2 \\
$\Vae_{80}$ & 484.5 $\pm$ 1.2 & 483.7 $\pm$ 1.2 \\
\hline
\end{tabular}
\caption{\small Estimated Negative Log-Likelihoods of \Test-25 when Sum-25 constraint is and is NOT satisfied by $q(z)$, for VAE conditional visual models trained for 10, 70 and 80 epochs. Notice how there is barely any improvement from having digits sum up to 25 or not (compare ``Sum-25 satisfied'' to ``NOT satisfied''), while even tiny improvements in visual quality yield substantial gains in NLL (compare $\Vae_{70}$,$\Vae_{80}$ which have no perceptible difference). \label{table:nllfactorized}}
\end{table}

\begin{figure}

\centering

\begin{subfigure}[t]{0.49\linewidth}
\centering
\includegraphics[width=\linewidth]{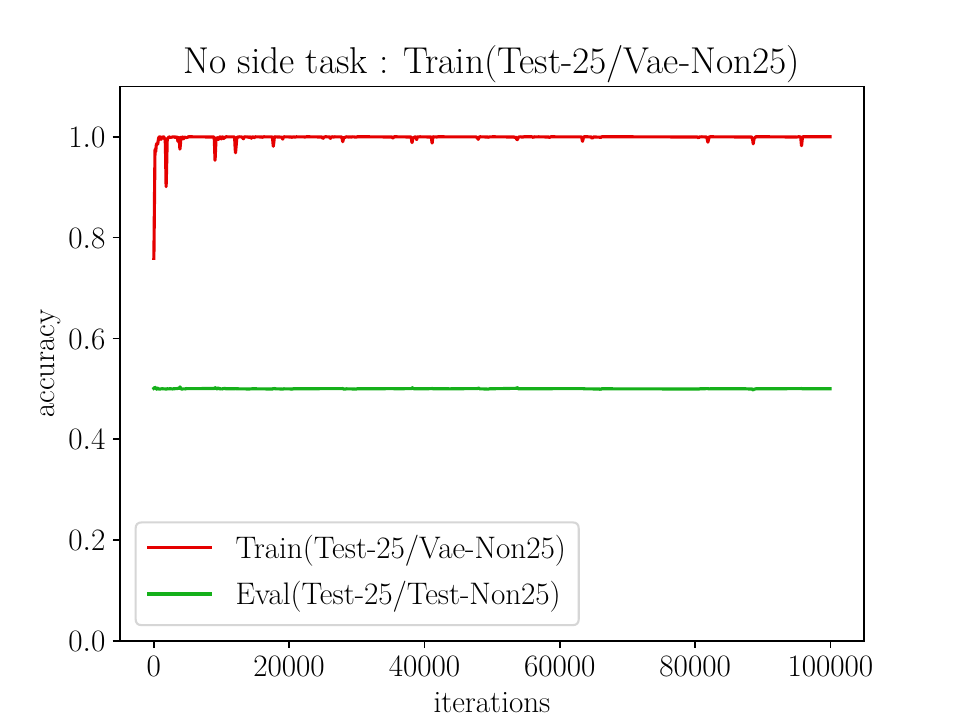}
\caption{\footnotesize Train:\Test-25/\Vae-Non25
} 
\end{subfigure}
\begin{subfigure}[t]{0.49\linewidth}
\centering
\includegraphics[width=\linewidth]{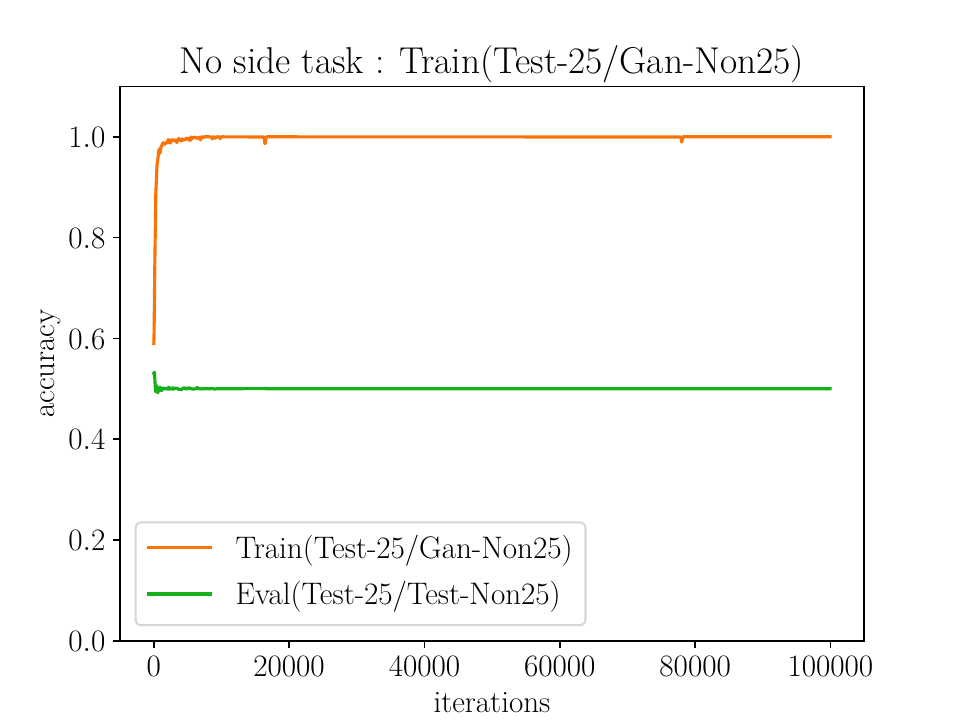}
\caption{\footnotesize Train: \Test-25/\Gan-Non25
} 
\end{subfigure}
\begin{subfigure}[t]{0.49\linewidth}
\centering
\includegraphics[width=\linewidth]{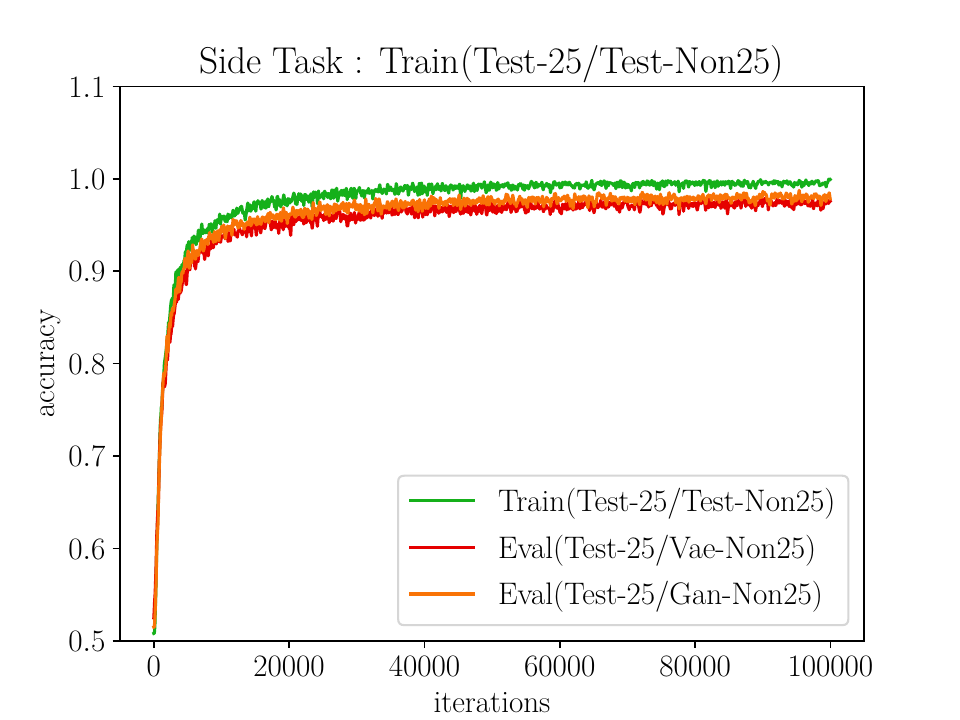}
\caption{\footnotesize Train: \Test-25/\Test-Non25
} 
\end{subfigure}

\caption{Probing the discriminator. Experiment \textbf{a} (resp \textbf{b}) : we train the discriminator to separate \Test-25/\Vae-Non25 (resp. \Test-25/\Gan-Non25)  training accuracy (red) goes to 1 very quickly. We probe the discriminator to see if it can separate  \Test-25/\Test-Non25, and observe bad accuracy (green), which means the discriminator was unable to learn the sum-25 constraint, and discriminated solely on visual properties. In Experiment \textbf{(c)}, we train the discriminator to separate \Test-25/\Test-Non25 as a side task. Good training accuracy (green) is evidence that the discriminator learned the sum-25 constraint, since it is the only way to separate the distributions. We then evaluate if the discriminator can separate \Test-25/\Vae-Non25 and get evaluation accuracy (red) almost as good training accuracy. Slightly higher evaluation accuracy is observed for \Test-25/\Gan-Non25, which makes sense since \Gan\  digits are visually more similar to \Test. Thus the discriminator can detect the sum-25 constraint despite the \Test~$\rightarrow$~\Vae\ and  \Test~$\rightarrow$~\Gan\ domain shifts, which suggests such the SideTask could help enforce Sum-25 during end-to-end training (i.e. training a GAN).\label{fig:probe}}

\end{figure}

\paragraph{Sensitivity of Parametric Divergence (a,b)} 

For our second experiment, we would like to compute the parametric Jensen-Shannon between the reference distribution \Test-25 and candidate distributions based on various conditional and symbolic models. We consider \Vae-25, \Vae-Non25, \Gan-25 and \Gan-Non25, where \Vae is a VAE trained for 80 epochs, and \Gan is a WGAN-GP trained for 50 epochs. 
However, we cannot present a similar table as previously because the numerical values of parametric divergences are unstable/random and dependent both on discriminator initialization and random sampling of data (even with gradient-penalty/other formulations). Instead, we propose an alternative approach which we call \textit{probing} the discriminator. Specifically, we first train the discriminator to classify \Test-25/\Vae-Non25 (resp. \Test-25/\Gan-Non25), but we evaluate its accuracy on the different problem of separating \Test-25/\Test-Non25 to see whether the discriminator has learned to detect the sum-25 constraint. As shown in plot \textbf{(a)} of Figure~\ref{fig:probe}, the discriminator fails to discriminate \Test-25/\Test-Non25, which suggests that it has only focused on the visual constraint during training. This is actually not that surprising as the discriminator can actually get excellent accuracy just by examining the visual quality of the first digit.

\paragraph{Sensitivity of Parametric Divergence w/ Side-Task (c)} In our third experiment, we explore whether we can better enforce the Sum25 constraint through the discriminator. For this purpose, we train the discriminator on the side-task of separating \Test-25/\Test-Non25. Since the individual digits are now indistinguishable, the only way for the discriminator to get maximum accuracy on the side-task is to consider digits jointly and detect if they sum to 25. After training the discriminator on the side task, we probe the discriminator to see whether it can separate \Test-25/\Vae-Non25 (resp. \Test-25/\Gan-Non25) which simulates the target and (imperfect) model distribution one could encounter while training a generative model. As shown in plot \textbf{(c)} of Figure~\ref{fig:probe}, it turns out the discriminator can separate \Test-25/\Vae-Non25 (resp. \Test-25/\Gan-Non25) fairly well, and moreover, it can also generalize to new combinations it has not seen before. This suggests that the side-task approach can help enforce sum-25 when learning $q$ end-to-end, in a GAN setting.

\subsection{Generating the Sum-25 Distribution \label{sec:25e2e}} 

The previous section suggested that KL-divergence does not care about the symbolic constraint, that the vanilla parametric divergence does not either, but the parametric divergence + side-task cares about the constraint. We investigate if our intuitions hold when actually training a generator using these divergences. We train three models to generate the training set: a \textbf{VAE} (objective is upper bound on nonparametric Kullback-Leibler), a WGAN-GP (objective is parametric Wasserstein, simply denoted \textbf{GAN}), and another WGAN-GP with the additional side-task of discriminating \Test-25/\Test-Non25 (objective is parametric Wasserstein with side task, denoted \textbf{GAN-SideTask}). All models share the same architecture for their generator network and use 200 latent variables. 
After training, with the help of a MNIST classifier, we automatically recognize and sum up the digits in each generated sample. As usual, the VAE samples are a bit blurry while the GAN samples are more realistic and crisp. Generated samples can be found in Section~\ref{sec:morevisual} of the appendix.  We then compare how well VAE, GAN and GAN-SideTask \emph{enforce} and \emph{generalize} the constraint that the digits sum to 25.

\begin{figure*}%
\centering
\includegraphics[width=0.49\linewidth]{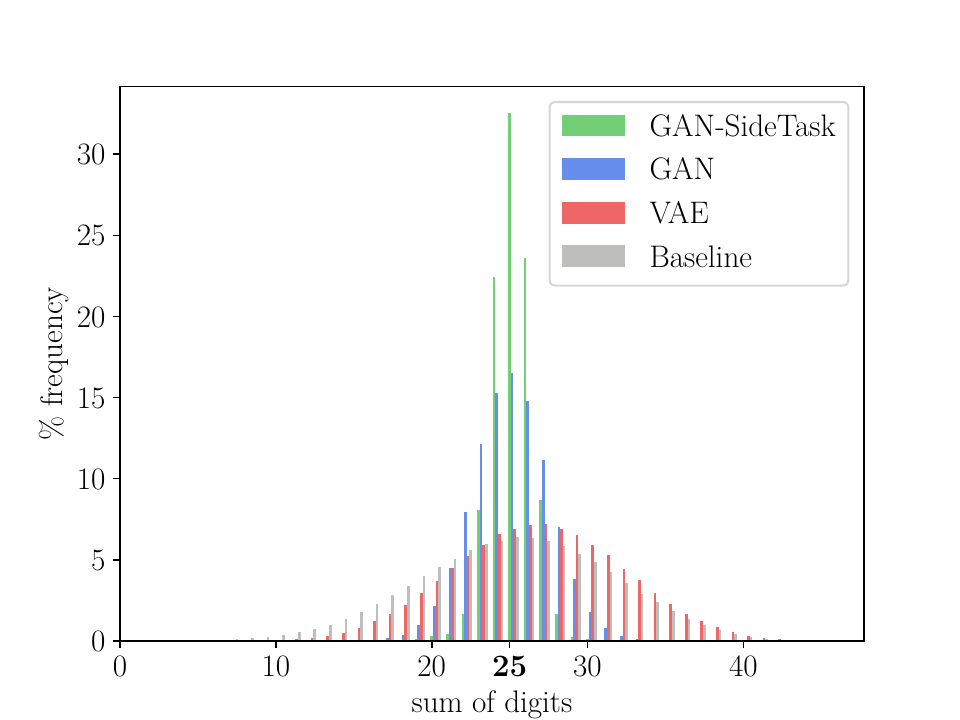}
\includegraphics[width=0.49\linewidth]{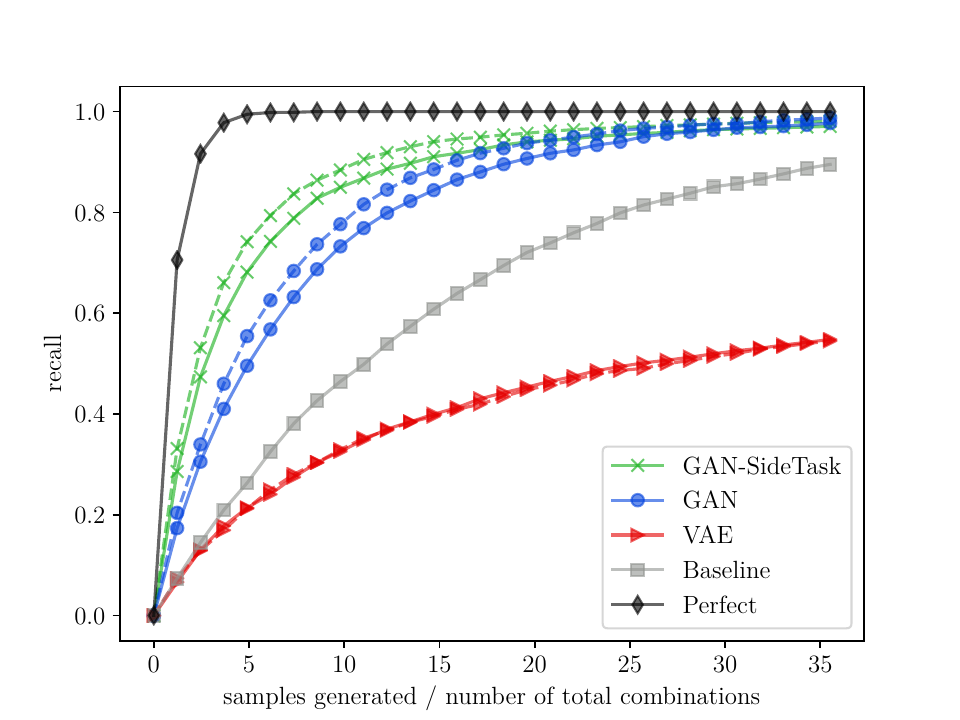}
\caption{\label{fig:visual-hyperplane} \small 
\textbf{Left:} Histograms of the sums of digits generated by VAE (red), WGAN-GP (green) and Independent Baseline (gray). The baseline draws digits independently according to their empirical marginal probabilities, which corresponds to fitting independent multinomial distributions over digits using maximum likelihood. WGAN-GP beats largely both VAE and Indepedent Baseline as it gives a sharper distribution centered on the target sum 25. 
\textbf{Right:} Train (dashed) and test (solid) which tell us how well the models cover the train and test set. The best theoretical recall is given by the Perfect generator (black) which samples uniformly among the 5631 combinations. Higher train recall means \emph{less mode dropping}, while higher test recall means \emph{better ability to generalize constraints}. WGAN-GP has the best recall (green), followed by the independent baseline (gray) and the VAE (red). Plots averaged over 5 runs. Train (dashed) and test (solid) recalls are identical for independent baseline and perfect generator.
}
\end{figure*}

\paragraph{Enforcing the Symbolic Constraint [Figure~\ref{fig:visual-hyperplane} left].} We plot the distributions of the sums of the combinations generated by the three models. The GAN-SideTask samples are the most concentrated around the target 25, followed by the GAN, and then VAE which is barely better than the independent baseline (digits follow marginal distribution). In that respect, the GAN-SideTask (though still far from nailing the problem) was much better than the VAE at capturing and enforcing the particular aspects and constraints of the data distribution (summing up to 25). This corroborates our prior hypothesis that side-task can help better align parametric divergences with the final task. Surprisingly, the GAN without side-task also outperforms the VAE. One hypothesis is that when GAN samples become visually too hard to discriminate, the discriminator will eventually start to detect the sum-25 constraint, even if such effect was not observed during fixed distribution experiments. Another possibility is that since the generator is always adapting to the discriminator (joint training), the discriminator does not have time to focus too much on visual details.

\paragraph{Generalizing the Symbolic Constraint [Figure~\ref{fig:visual-hyperplane} right].} We plot the train and test recall scores, which are defined as the proportions of the train/test combinations covered by a generative model after generating a fixed number of samples. High \textit{train} recall means the generators cover the training combinations well (no symbolic \textbf{mode dropping}), while high \textit{test} recall means they are able to generate new combinations which were not in the training set (symbolic \textbf{generalization}). Again, GAN-SideTask has best train/test recalls, followed by GAN, and VAE ranks last with same recall as the Independent Baseline.  There is a very slight overfitting to the training set for the GANs (solid line lower than dashed line), but GAN-SideTask seems to generalize very well. For models with very low recalls such as VAE, train and test recalls are equal, which is not surprising (it is the case for the uniform distribution). We leave for future work to investigate whether the generalization of the constraint is due more to the discriminator or the generator.

\paragraph{Conclusion of Sum-25.} To conclude the Sum-25 experiments, parametric divergences have allowed us, through the addition of a side-task, to better enforce the symbolic constraint, and thus yield generative models which better solve the Sum-25 task than ones trained with nonparametric divergences or vanilla parametric divergences. We can imagine that for more complex tasks, practitioners can create side-tasks to make the discriminators more sensitive to certain aspects of the distribution.

\section{Interactions between Generator and Divergence \label{sec:interactions}}

Previously in Sections~\ref{sec:scale}~and~\ref{sec:reflecttask}, we gave arguments against using f-divergences for general generators because f-divergences cannot directly handle implicit generators or be tuned to reflect the final task. Here, we discuss the fact that certain generators with a special structure can compensate for the shortcomings of the KL-divergence, but this special structure can also bring other problems (Section~\ref{sec:interactionskl}). We also consider the \textit{converse} question of whether we can train a memorization-based generators with no generalization ability, but using a parametric divergence (Section~\ref{sec:memorization}).

\subsection{Interaction of KL-divergence with Special Generators\label{sec:interactionskl}}

Certain generators with a special structure can compensate for the shortcomings of the KL-divergence. Here, the two special structures we discuss are : smoothing the generator distribution with a (Gaussian) observation model and autogressive models.

\paragraph{Smoothing observation model.} 

By adding an observation model such as a Gaussian model, on top of any generator, one can artificially extend its support to the whole input space. In particular, this makes the KL-divergence well-defined, and makes it possible to train models such as variational autoencoders\citep{kingma2013auto} (VAEs). The observation model makes the log-likelihood involve a ``reconstruction loss'', a pixel-wise $L_2$ distance between images analogous to the Hamming loss, which makes the training relatively easy and very stable. However, the Gaussian is partly responsible for the VAE's inability to learn sharp distributions. Indeed it is a known problem that VAEs produce blurry samples~\citep{arjovsky2017wasserstein}, and in fact even if the approximate posterior matches exactly the true posterior, which would correspond to the evidence lower-bound being tight, the output of the VAE would still be blurry~\citep{bousquet2017optimal}.

\paragraph{Autoregressive models.} Another example of special structure is autoregressive models, such as recurrent neural networks~\citep{mikolov2010recurrent}, which factorize naturally as $q_\theta(x) = \prod_i q_\theta(x_i|x_1, .., x_{i-1})$, and PixelCNNs~\citep{oord2016pixel}. Training autoregressive models using maximum likelihood results in teacher-forcing~\citep{lamb2016professor}: each ground-truth symbol is fed to the RNN, which then has to maximize the likelihood of the next symbol. Since teacher-forcing induces a lot of supervision, it is possible to learn using maximum-likelihood. Once again, there are similarities with the Hamming loss because each predicted symbol is compared with its associated ground truth symbol. However, among other problems, there is a discrepancy between training and generation. Sampling from $q_\theta$ would require iteratively sampling each symbol and feeding it back to the RNN, giving the potential to accumulate errors, which is not something that is accounted for during training. See~\citet{leblond2017searnn} and references therein for more principled approaches to sequence prediction with autoregressive models.

\subsection{Ability to Train Memorization-based Generators (Experiment)\label{sec:memorization}}

Previously, we discussed using generators with special structure to compensate shortcomings of the KL-divergence. Here we explore the converse case. Can we train a memorization-based generator, which has no generalization abilities, using a parametric divergence? Obviously, we cannot expect the generator to do any generalization, but this experiment is a good sanity check to see whether a given divergence will enforce realistic samples. Additionally, the memorized data can be plotted as a summary of the target distribution.

We compare the parametric Wasserstein divergences induced by three different discriminators (linear, dense, and CNN) under the WGAN-GP~\citep{gulrajani2017improved} formulation. The memorization-based generator is a mixture of 100 prototypes, which can be also thought of as a mixture of 100 Gaussians with zero-variance. To sample a new image, the generator randomly returns one of 100 learned images. The model ``density'' is $q_\theta(\x) = \frac 1 K \sum_z \delta(\x-\x_z)$, where $x_z$ are the prototypes (images) and $\delta$ is the Dirac distribution.

\begin{figure}
\centering
\includegraphics[width=0.32\linewidth]{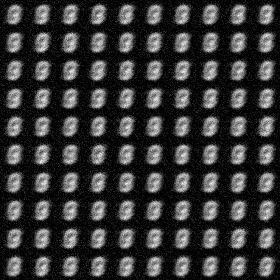}
\includegraphics[width=0.32\linewidth]{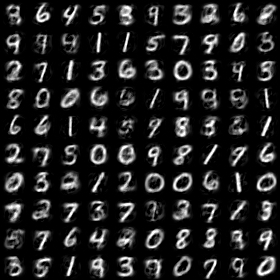}
\includegraphics[width=0.32\linewidth]{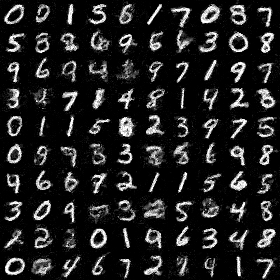}
\caption{All 100 Prototypes learned using linear (\textbf{left}), dense (\textbf{middle}), and CNN discriminator (\textbf{right}). We observe that with the linear discriminator, only the mean of the training set is learned, while using the dense discriminator yields blurry prototypes. Only the CNN discriminator yields clear prototypes.  \label{fig:mixdirac-more}}
\end{figure}

Prototypes learned from MNIST are shown in \textbf{Figure~\ref{fig:mixdirac-more}}. The first observation is that the linear discriminator is too weak of a divergence: all prototypes only learn the mean of the training set. Now, the dense discriminator learns prototypes which sometimes look like digits, but are blurry or unrecognizable most the time. The samples from the CNN discriminator are never blurry and recognizable in the majority of cases. Our results confirms that indeed, even for simplistic models like a mixture of Diracs, using a CNN discriminator provides a better task loss for generative modeling of images. 

Note that it would have been impossible to directly fit the same generator using a KL-divergence, because such a model does not admit a continuous density. A workaround approach is to relax the generator into a model that admits a density with respect to the usual measure, by replacing the Diracs with Gaussians. Maximizing the likelihood could then be done using an algorithm similar to K-means, and would likely result in blurry prototypes, since each prototype would be expressed as the average of several images.

\section{Parametric Divergences for Meaningful Mutual Information\label{sec:pmi}}

Recall that the mutual information between two variables $X$ and $Y$ can be defined as the Kullback-Leibler divergence between the joint distribution and the product of marginals\footnote{We use densities for simplicity, but more general definitions based on measure theory exist.}
\begin{align}
\mathbf{MI}(X,Y)=\int_{x,y} p(x,y) \log \frac{p(x,y)}{p(x)p(y)} dxdy = \textbf{KL}(p(x,y) || p(x)p(y) )
\end{align}
Mutual information quantifies the amount of information, in \textit{nats} or \textit{bits}, which is shared between $X$ and $Y$, independently of the way these variables are represented. 

We argue that the problems of the KL-divergence presented previously can lead to problematic properties of mutual information. We illustrate that mutual information is not always an intuitive concept and can have a paradoxical behavior (Sections~\ref{sec:permutedmnist}~and~\ref{sec:hashingparadox}). We give an explanation of these paradoxes in Section~\ref{sec:resolveparadox}, and propose more intuitive and meaningful notions of generalized mutual information based on parametric divergences in Section~\ref{sec:pmidef}. Finally, we apply the proposed generalized mutual information to the paradoxes and discuss the results.

\subsection{The Corrupted-Label Paradox\label{sec:permutedmnist}}

We showcase a simple example of joint distribution $p(x,y)$ where the mutual information is high between $x$ and $y$ even if there is no meaningful dependency between them.

Consider $\lbrace (x_i,y_i) \rbrace_{1\leq i\leq N}$ the labeled training examples of MNIST (although the paradox can be derived for any classification problem). There are $N=60,000$ examples and $K=10$ balanced classes, each class containing exactly $N/K=6,000$ examples. Consider two joint distributions for which we compute the mutual information using the difference between marginal and conditional entropy $\mathbf{MI}(X,Y)=H(Y)-H(Y|X)$:
\begin{itemize}
\item \textbf{Empirical}: Define $p_1(x,y)$ as the empirical distribution of the training set $p_1(x,y)=\frac{1}{N}\sum_{i=1}^N \delta(x-x_i) 1(y=y_i)$, where $\delta(\cdot)$ is the Dirac distribution on $\mathcal X$. The marginal distribution of $Y$ is uniform over 10 classes, so the marginal entropy of $Y$ is $H(Y)=\log 10=2.30$ nats. For the conditional entropy, when $x$ is known and $p_1(x)>0$, the label is fully determined by $y=\sum_{i=1}^N 1(x=x_i) y_i$, so $H(Y|X)=0$ nats, and the mutual information is $\mathbf{MI}(X,Y)=2.3$ nats. 
\item \textbf{Pseudo-Random labels}: 
Sample and fix random permutation $\sigma(i)$ over $60,000$ elements, and permute all labels by considering the distribution $p_2(x,y)=\sum_{i=1}^N \delta(x=x_i) \delta(y=y_{\sigma(i)})$.  The marginal distribution of $Y$ does not change, so $H(Y)=\log 10=2.30$ nats again. For the conditional entropy, the label is now fully determined\footnote{Assuming no duplicate images.} by $y=\sum_{i=1}^N 1(x=x_i) y_{\sigma(i)}$, so $H(Y|X)=0$ nats, and the mutual information is $\mathbf{MI}(X,Y)=2.3$ nats again (assuming fixed $\sigma$).
\end{itemize}

It can be rather surprising that the mutual information in both cases is always approximately $2.3$ nats, which corresponds the amount of information for disambiguating between 10 balanced classes. While the labels for the empirical distribution are very natural and correspond to the identity of the digit represented, the labels for the permuted-label distribution have no meaning at all and would appear random to any human. Clearly, the mutual information does not care whether $X$ and $Y$ having a \textit{meaningful} dependency. Instead, it will always equal $2.3$ nats as long as the same label is consistently assigned to each image.

\subsection{The Hashing Paradox\label{sec:hashingparadox}}

We showcase a family of distributions $p_K(X,Y)$ -- this time over continuous variables $X,Y\in\R$-- for which the mutual information $\mathbf{MI}(X,Y)$ goes to infinity, while $X$ and $Y$ appear to be increasingly independent. Recall that continuous mutual information is always non-negative and independent of the base-measure, and preserves the meaning that $\mathbf{MI}(X,Y)=0$ if and only if $X$ and $Y$ are independent. If $Y=X$, then the mutual information equals infinity.

For any positive $K\in \mathbb{N}$, define the distribution $p_K(X,Y)$ as follows. Consider two independent uniform random variables $X,W \sim_{iid} U\left([0,1)\right)$, and a (deterministic) permutation $\sigma_K$ of the range $\lbrace 0, \dots, K-1 \rbrace$, which we call the \textit{hash} function and will be defined later.
We define the random variable $Y=\frac{\sigma_K(\textbf{floor}(K*X))}{K} + \frac{W}{K}$. It is very easy to verify that the marginal distribution of $Y$ is also uniform on $[0,1)$, so its marginal differential entropy is $H(Y)=0$. For the conditional entropy, when $X$ is known, we know that $Y$ is uniform in an interval of measure $1/K$, so $H(Y|X)=\log \frac{1}{K} = -\log K$, regardless of the actual permutation $\sigma_K$. Therefore, the mutual information is always $\log K$ nats, and grows to infinity as the number of bins goes to infinity.

\begin{figure}
\includegraphics[width=\textwidth]{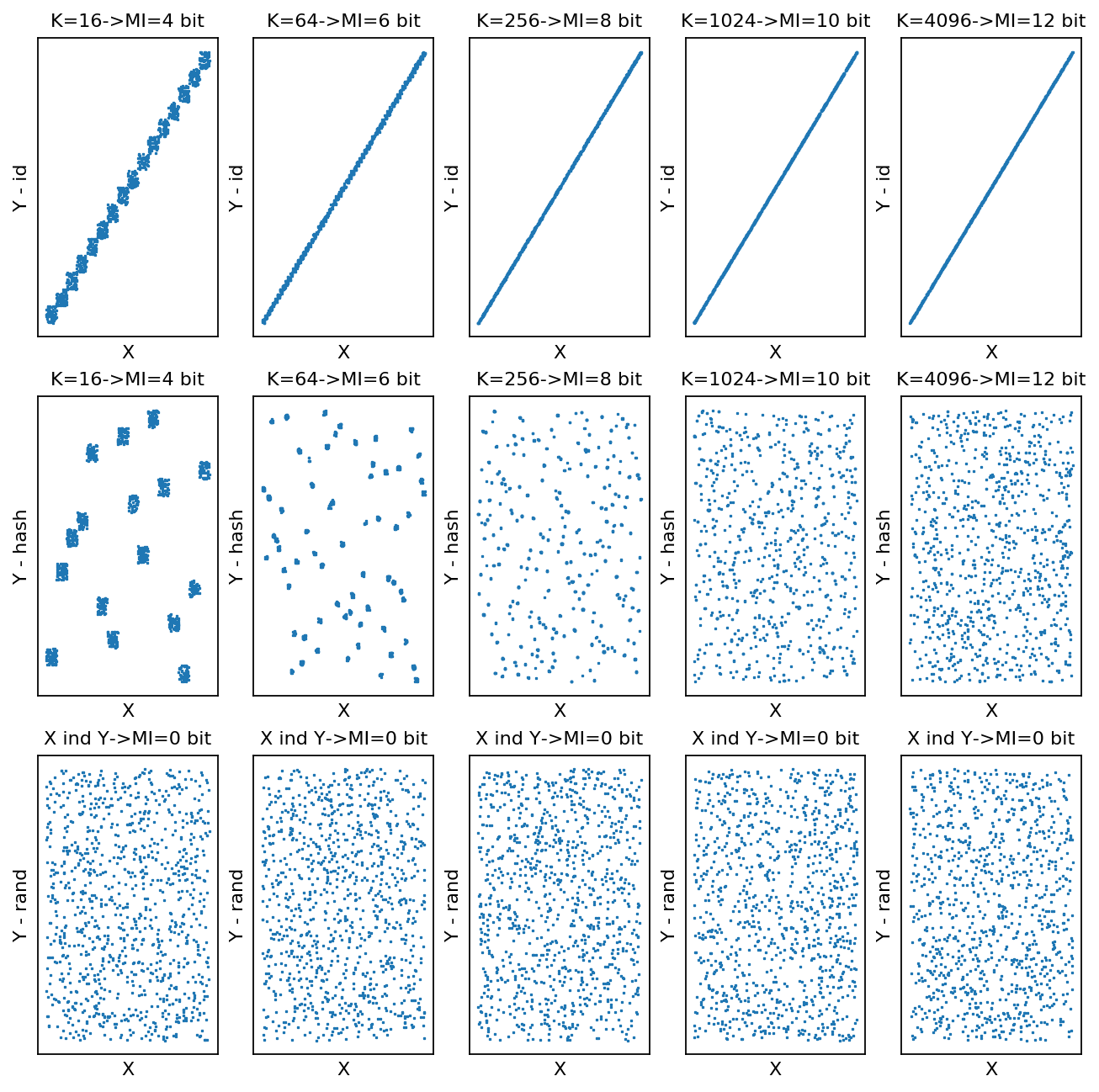}
\caption{The Hashing Paradox. Samples of the distribution $p_K(x,y)$ for increasing numbers of bins $K\in\lbrace 16,64,256,1024,4096 \rbrace$, with corresponding mutual information. \textbf{Row 1}: the permutation $\sigma_K$ is the identity. \textbf{Row 2}: the permutation is a pseudo-random permutation. \textbf{Row 3}: $X$ and $Y$ are sampled independently. Observe how the samples from Row 2 appear to be more and more independent (and distributed similarly to Row 3), despite having increasingly higher mutual information. In fact, for high enough $K$, it is impossible to tell Row 2 from Row 3.\label{fig:hashingparadox}}
\end{figure}

We consider two families of hash functions $\sigma_K$. For increasing $K$, we represent samples $(X,Y)\sim p_K$ along with the mutual information $\mathbf{MI}(X,Y)$ \textbf{[Figure~\ref{fig:hashingparadox}]}:
\begin{itemize}
\item \textbf{Identity function} $\sigma_K(i)=i$. As $K$ grows, $Y$ has to be closer and closer to $X$, while their mutual information goes to infinity. This is intuitive because $Y$ is essentially converging towards $X$ and giving more and more information over $X$.
\item \textbf{Pseudo-random permutation}. Consider any given implementation of Python. Seed the pseudo-random number generator (PRNG) to 0 (or any fixed number) and define a permutation $\sigma$ by shuffling an array containing the range $\lbrace 1,\dots, K\rbrace$. As shown in Figure~\ref{fig:hashingparadox}, when $K$ grows to infinity, $Y$ looks visually more and more independent from $X$, thus we would intuitively expect the mutual information to vanish to zero. However, as proved previously, the mutual information does not depend on $\sigma$, and actually grows in $\log K$, which \textit{contradicts} (in appearance) the fact that $X$ and $Y$ are \textit{visually} more and more independent. Beyond visual independence, it would be easy to show that for large enough $K$, the random variables $X$ and $Y$ can be considered numerically independent for practical purposes, such as numerical integration $\int_{x=0}^1 \int_{y=0}^1  f(x,y) dp_K(x,y)$, as long as the function $f$ is sufficiently smooth.
\end{itemize}

\subsection{Statistical Dependency can be Arbitrarily Complex \label{sec:resolveparadox}}

The previous paradoxes comes from the fact that mutual information does not care about any underlying \textit{metric} or \textit{similarity} on the spaces $\mathcal X$ and $\mathcal Y$. Two points are either considered equal or not equal, which is a property that stems from the KL divergence. Moreover, the mutual information also does not care about the nature of the relationship between $X$ and $Y$. Indeed, no matter how ``complex'' the relationship is, as long as there exists some function such that $Y=h(X)$, then all of $Y$ is determined by $X$, and the mutual information between $Y$ and $X$ is maximal, i.e., equal to the entropy of $Y$.

If the relationship between $X$ and $Y$ is too \textit{complex}, i.e., it is too \textit{hard} to predict $Y$ from $X$, then can we really consider that $X$ and $Y$ are dependent for any practical purposes? Consider the hashing example, where $Y$ appears to be visually random from $X$. For applications like numerical integration of smooth functions, it is likely that we can expect the same type of guarantees as if $X$ and $Y$ were truly independent. However, for cryptographic applications where randomness requirements are much more stringent, we can probably not consider that $X$ and $Y$ are independent. In fact, we could even push the reasoning one step further and argue that pseudo-random number generators are not truly random. However, PRNGs such as linear congruential generators (LCG) are omnipresent in machine learning (e.g. SGD sampling, network initialization, sampling latent variables) and generally accepted to be truly random for all practical purposes, although successive samples are known to be highly dependent in a strict sense.

These apparent paradoxes motivate the need for defining more \textit{meaningful} notions of mutual information, which are tailored to the specificities of the final application (e.g. numerical integration, sampling, assessing independence, learning disentangled latent variables).
 One way to do this is to control the complexity of the relationship $Y=h(X,\epsilon)$ between $X$ and $Y$, where $h$ is deterministic and $\epsilon$ is some noise independent from $X$. For instance, we can restrict the class of possible relationships $h\in\mathcal H$, or define a Bayesian prior $p(h)$ over them:
\begin{itemize}
\item \textbf{Linear Dependence.} If only linear dependence is considered simple, then the resulting notion of independence corresponds to the usual notions of linear correlation.
\item \textbf{Smoothness.} Define some metric on $\mathcal X$ and some metric $\mathcal Y$. We could restrict $\mathcal H$ to the class of $L$-Lipschitz functions for some $L$, which means only smooth relationships are deemed interpretable.
\item \textbf{Neural Networks.} Consider the set of neural networks with a fixed neural network architecture. We could restrict $h$ to be representable with one of these neural networks. Alternatively, we could also restrict $h$ to be a neural network which can be fitted within a limited number of SGD steps, in order to take advantage of its implicit regularization properties.
\item \textbf{Kolmogorov complexity.} Given a programming language, we could theoretically define the complexity of a function as the length of the shortest program for implementing that function. For instance, this language could be Python, restricted to its standard library. 
\item \textbf{Arbitrary Complexity / Memorization-based} If we consider all the functions $h\in\mathcal H=\mathcal Y^{\mathcal X}$, then any arbitrary pair $(X,Y)$ can just be \textit{memorized} inside $h$. 
By definition, any $h$ is considered simple in the memorization-sense.
In this case, we recover the classic definition of mutual information, which allows $h$ to be arbitrarily complex.
\end{itemize}
We will be introducing complexity-aware variants of mutual information in Section~\ref{sec:pmidef}, and experimenting with some of their properties in Section~\ref{sec:semidiscrete}~and~\ref{sec:miexp}.

\paragraph{Explaining the corrupted label paradox. }
In the corrupted-label paradox for the random label case, each label $Y$ is assigned according to an arbitrary rule $\sigma$, and cannot be deduced from the image $X$ other than through memorization. Here the relationship $Y=h(X)$ is complex in the smoothness, neural network, and Kolmogorov sense, and only simple for the memorization-based sense. Therefore, if complexity in the sense of smoothness, neural network, and Kolmogorov complexity are meaningful to the final task (e.g. identifying semantic correlation), then an intuitive notion of mutual information should predict $X$ and $Y$ to be \textit{independent}.

\paragraph{Explaining the Hashing Paradox} 
In the hashing paradox, when $\sigma_K=hash$ is a pseudo-random permutation, the relationship $h_K$ between $X$ and $Y$ is \textit{complex} in the smoothness and neural network sense, but \textit{simple} in the Kolmogorov sense because $h_K$ can be implemented in a few lines of python. Therefore, if complexity in the sense of smoothness and neural networks are meaningful to the application (e.g. numerical integration of smooth functions), we should consider the variables to be \textit{independent}. However, if complexity in the sense of Kolmogorov are meaningful to the application (e.g. cryptography), then the variables \textit{cannot} be considered independent.

\subsection{Generalized and Parametric Mutual Information\label{sec:pmidef}}

We explain how to generalize mutual information to account for arbitrary properties of the distribution, such as the complexity of the relationship $Y=h(X)$ and underlying metrics of the spaces $X,Y$. Recall that the mutual information $\mathbf{MI}(X,Y)$ between two random variables $X$ and $Y$ can be written as the KL-divergence between the joint $p_{X,Y}$ and the product of marginals $p_X\otimes p_Y$.
\begin{align}
\mathbf{MI}(X,Y) &= \mathbf{KL}(p_{X,Y} || p_X\otimes p_Y ) 
\end{align}
We define the \textit{generalized mutual information} by replacing the KL-divergence with another nonparametric and parametric divergences.

\begin{definitiontwo}[Generalized Mutual Information]
Given a divergence $\mathbf{Div}(\cdot||\cdot)$, we define the $\mathbf{Div}$-generalized mutual information (GMI) as:
\begin{align}
\mathbf{GMI}_{\mathbf{Div}}(X,Y) &= \mathbf{Div}(p_{X,Y} || p_X\otimes p_Y ) 
\end{align}
Additionally, if $\mathbf{Div}(\cdot||\cdot)$ is a parametric divergence, then we say that $\mathbf{GMI}_{\mathbf{Div}}$ is a parametric mutual information (PMI).
\end{definitiontwo}

\begin{definitiontwo}[Generalized Independence]
Given a divergence $\mathbf{Div}(\cdot||\cdot)$, we say that $X$ and $Y$ are $\textbf{Div}$-independent if and only if $\mathbf{GMI}_{\mathbf{Div}}(X,Y)=0$. We abuse the terminology and say that $X$ and $Y$ are $\textbf{Div}$-independent even when $\textbf{Div}$ is not a proper divergence.
\end{definitiontwo}

For instance, we could consider the class of f-divergences (KL, Jensen-Shannon), integral probability metrics (MMD, Wasserstein) or parametric adversarial divergences. We argue that the properties of parametric divergences, discussed in Sections~\ref{sec:scale}~and~\ref{sec:reflecttask} transfer over to the induced parametric mutual information.

Using MMD for independence-testing has been proposed in~\citet{gretton2012kernel}. The kernel defines a similarity metric over $\mathcal X\times \mathcal Y$. It should be noted that MMD-generalized-mutual-information can only be as meaningful as the kernel considered. In particular, for generic kernels MMD-GMI might not be powerful enough to find some dependencies, for the same reasons MMD can fail to discriminate distributions in high dimensions (Section~\ref{sec:samplecomplexity}).
Similarly, Wasserstein distance and variants have been proposed for independence-testing~\citep{ramdas2017wasserstein}. Just like for MMD, the choice of base-metric determines the properties of the generalize mutual information. However, similarly to the Wasserstein distance, Wasserstein-GMI also suffers from poor sample complexity in high dimensions.
KL-parametric mutual information (KL-PMI) has been proposed in~\citep{belghazi2018mine} under the name of Mutual Information Neural Estimator (MINE), as an approximator of the true mutual information. We argue that KL-parametric mutual information should not be seen as a mere approximator of mutual information, and that it can sometimes be a more natural and meaningful concept than traditional mutual information. We provide evidence of this in the experimental section~\ref{sec:miexp}.

\paragraph{Training and Validation GMI.}

In practice, generalized mutual information is estimated from a finite dataset, and it is handy to lower and upper bound the population generalized mutual information. We independently sample a training set $D_{train}=\lbrace (\x_{train}^{(i)}, \y_{train}^{(i)}) \rbrace$ and a validation set $D_{val}=\lbrace (\x_{val}^{(i)}, \y_{val}^{(i)}) \rbrace$ from the joint distribution $p_{X,Y}$.
\begin{definitiontwo}[Training GMI] 
Denote $N$ the size of the training set. We define the \textit{training} GMI as:
\begin{align} \label{eq:def-training-gmi}
\textbf{GMI}(X,Y) _{train} =  \sup_{f\in\F} \frac{1}{N(N-1)(N-2)} \sum_{i=1}^N \sum_{j\neq i}^N  \sum_{k\neq i,j}^N  \Delta( f(\x_{train}^{(i)}, \y_{train}^{(i)}),  f(\x_{train}^{(j)}, \y_{train}^{(k)}))
\end{align}
\end{definitiontwo}
In a similar spirit as the U-statistics-based MMD estimator proposed in Lemma 6 of~\citet{gretton2012kernel}, we remove some indices $j\neq i$ and $k\neq i,j$ so that we can reuse samples from the joint $p_{X,Y}$ as if $(\x_{train}^{(i)}, \y_{train}^{(i)}),(\x_{train}^{(j)}, \y_{train}^{(k)})$ were s   ampled from $p_{X,Y}\otimes (p_{X}\otimes p_Y)$. The resulting term inside the $\sup_{f\in\F}$ is an unbiased estimator of the term inside the $\sup_{f\in\F}$ of the corresponding adversarial divergence:
\begin{align}
&\mathbf E_{D_{train}} \left[ \frac{1}{N(N-1)(N-2)} \sum_{i=1}^N \sum_{j\neq i}^N  \sum_{k\neq i,j}^N  \Delta( f(\x_{train}^{(i)}, \y_{train}^{(i)}),  f(\x_{train}^{(j)}, \y_{train}^{(k)})) \right]
\\= &\underbrace{\mathbf E_{(x,y),(x',y') \sim p_{X,Y} \otimes (p_{X}\otimes p_Y)} \left[ \Delta(f(x,y),f(x',y') \right]}_{\text{take $\sup_{f\in\F}$ to get $GMI(X,Y)$}}
\end{align}
However, it is important to note that~\eqref{eq:def-training-gmi} is not an unbiased estimator of the GMI because of the supremum. We will only derive a bound in expectation in Theorem~\ref{theorem:bounds-gmi}.

\begin{definitiontwo}[Validation GMI]
Denote $\widehat f$ the previous maximizer, and $N'$ the size of the validation set. We define the \textit{validation} GMI as:
\begin{align}
\textbf{GMI}(X,Y) _{val} =  \frac{1}{N'(N'-1)(N'-2)} \sum_{i=1}^{N'} \sum_{j\neq i}^{N'}  \sum_{k\neq i,j}^{N'}  \Delta(\widehat f(\x_{val}^{(i)}, \y_{val}^{(i)}), \widehat f(\x_{val}^{(j)}, \y_{val}^{(k)})).
\end{align}
\end{definitiontwo}
Using the same principles as for supervised classification (Section~\ref{sec:practice}) the true (population) GMI can be lower (resp. upper) bounded by the training (resp. validation) GMI~\eqref{eq:pmibounds}. 

\begin{theoremtwo}[Bounds for Generalized Mutual Information]\label{theorem:bounds-gmi}
For any generalized mutual information, the following bounds hold:
\begin{align}
\mathbf E_{D_{train},D_{val}} [{\mathbf{GMI}}_{val}(X,Y) ]\leq \mathbf{GMI}(X,Y) \leq \mathbf E_{D_{train}} [{\mathbf{GMI}}_{train}(X,Y)]. \label{eq:pmibounds}
\end{align}
\end{theoremtwo}

\begin{proof} For conciseness, denote:
\begin{align}
    \widehat {\Delta}_{train} = \frac{1}{N(N-1)(N-2)} \sum_{i=1}^N \sum_{j\neq i}^N  \sum_{k\neq i,j}^N  \Delta( f(\x_{train}^{(i)}, \y_{train}^{(i)}),  f(\x_{train}^{(j)}, \y_{train}^{(k)})).
\end{align}
The right inequality results from the fact that taking the supremum inside the expectation $\E_{D_{train}}[\cdot]$ is always higher than outside it, and the fact that $\widehat {\Delta}_{train}$ is an unbiased estimator of $\mathbf{GMI}(X,Y)$ without the supremum:
\begin{align}
 \mathbf{GMI}(X,Y) = \sup_{f\in\F} \underbrace{ \E_{D_{train}}[\widehat {\Delta}_{train}]}_{\mathbf E_{(x,y),(x',y') \sim p_{X,Y} \otimes (p_{X}\otimes p_Y)} \left[ \Delta(f(x,y),f(x',y') \right]} \leq  \E_{D_{train}}[\underbrace{ \sup_{f\in\F} \widehat {\Delta}_{train}}_{{\mathbf{GMI}}_{train}(X,Y)} ].
\end{align}
For the left inequality, taking the expectation of $\mathbf{GMI}(X,Y)_{val}$ with respect to the sampling of the validation set gives:
\begin{align}
\E_{D_{val}} [ \mathbf{GMI}(X,Y)_{val} ] =&
{\mathbf E_{(x,y),(x',y') \sim p_{X,Y} \otimes (p_{X}\otimes p_Y)} \left[ \Delta(\widehat f(x,y),\widehat f(x',y') \right]}
\\ \leq&  \sup_{f\in\F} \E_{(\x,\x')\sim p\otimes {q_\theta}}\left[\Delta( f(x,y), f(x',y') \right]
\end{align}
where the inequality comes from the definition of the supremum. Now, we take expectations with respect to the sampling of the training set, which $\widehat f$ depends on:\footnote{That last step is optional: the expectation with respect to $D_{train}$ could be removed.}
\begin{align}
\E_{D_{train},D_{val}} [\mathbf{GMI}(X,Y)_{val} ] \leq \mathbf{GMI}(X,Y)
\end{align}
\end{proof}

We will be using these bounds to estimate the GMI in the experiments of Section~\ref{sec:miexp}.

\subsection{MMD-Independence for Integrating Smooth Functions.}

We take the example of numerical integration of smooth functions to illustrate that weaker notions of independence can be sufficient for practical purposes. Specifically, consider a smooth function $f(X,Y)$, in the sense that $f\in RKHS$ and $||f||_{RKHS} < \infty$. We want to compute its integral $I(f)$ over $[0,1]^2$, which can also be written as an expectation by defining two \textit{independent} random variables $X,Y$ which are uniform over $[0,1]$ :
\begin{align}
I(f) &= \int_{x=0}^1 \int_{y=0}^1 f(x,y) dy dx = \int_{x\in X}\int_{y\in Y} f(x,y) p_Y(y) p_{X}(x) dy dx = <f, p_X\otimes p_Y>_{L2}
\end{align}
However, we can actually relax the independence assumption and only assume that $X$ and $Y$ are approximately MMD-independent, in the sense that their weak MMD-mutual information $\textbf{GMI}_{\textbf{MMD}}(X,Y)$ is small or equal to zero. Under that assumption, the new integral is 
\begin{align}
J(f) &= \int_{x\in X}\int_{y\in Y} f(x,y) p_{X,Y}(x,y) dy dx = <f, p_{X,Y}>_{L2}
\end{align}
To show that $J(f)$ approximates $I(f)$ better as $X$ and $Y$ become more and more weakly MMD-independent, we write
\begin{align}
|J(f)-I(f)| &= |<f, p_{X,Y} - p_X\otimes p_Y>_{L2}| = 
|<f,\mu_{p_{X,Y}} - \mu_{p_X\otimes p_Y}>_{RKHS}|
\end{align}
where the second equality comes from the definition of MMD embeddings $\mu_{p_{X,Y}},\mu_{p_X\otimes p_Y}$ and the fact that $f$ is in $RKHS$. Applying the Cauchy-Schwartz inequality and the definition of MMD-based mutual information yields
\begin{align}
|J(f)-I(f)|  \leq ||f||_{RKHS} * ||\mu_{p_{X,Y}} - \mu_{p_X\otimes p_Y}||_{RKHS}
\leq ||f||_{RKHS} * \underbrace{\textbf{GMI}_{\textbf{MMD}}(X,Y)}_{\text{small or zero}} \label{eq:jeqi}
\end{align}
Therefore, whether $X$ and $Y$ are strictly independent or only MMD-independent does not make any difference for integrating smooth functions in RKHS, because in both cases we will have $J(f)=I(f)$.

\paragraph{GMI for the Hashing Paradox.}  

Recall the distributions $p_{X,Y}$ introduced for the hashing paradox (Section~\ref{sec:hashingparadox}): 
\begin{itemize}
\item Identity case: $X\sim U([0,1])$ and $Y=\frac{\textbf{floor}(K*X)}{K} + \frac{W}{K}$. Variables $X$ and $Y$ appear more and more correlated, which is consistent with $\textbf{MI}(X,Y)=\log K \to_{K\to\infty} \infty$.
\item Hashing case: $X\sim U([0,1])$ and $Y=\frac{\textbf{hash}_K(\textbf{floor}(K*X))}{K} + \frac{W}{K}$. For large $K$, variables $X$ and $Y$ appear visually more and more independent, which seems incompatible with $\textbf{MI}(X,Y)=\log K \to_{K\to\infty} \infty$.
\item Independent case: $X\sim U([0,1])$ and $Y\sim U([0,1])$ are independent and $\textbf{MI}(X,Y)=0$, which is consistent with the visual appearance.
\end{itemize}
\begin{figure}
\centering
\includegraphics[width=0.48\textwidth]{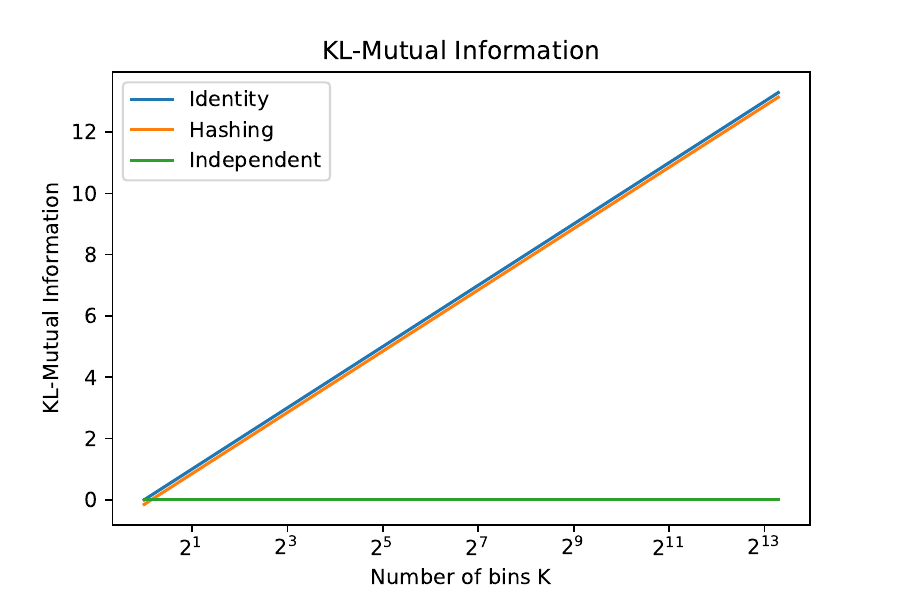}
\includegraphics[width=0.48\textwidth]{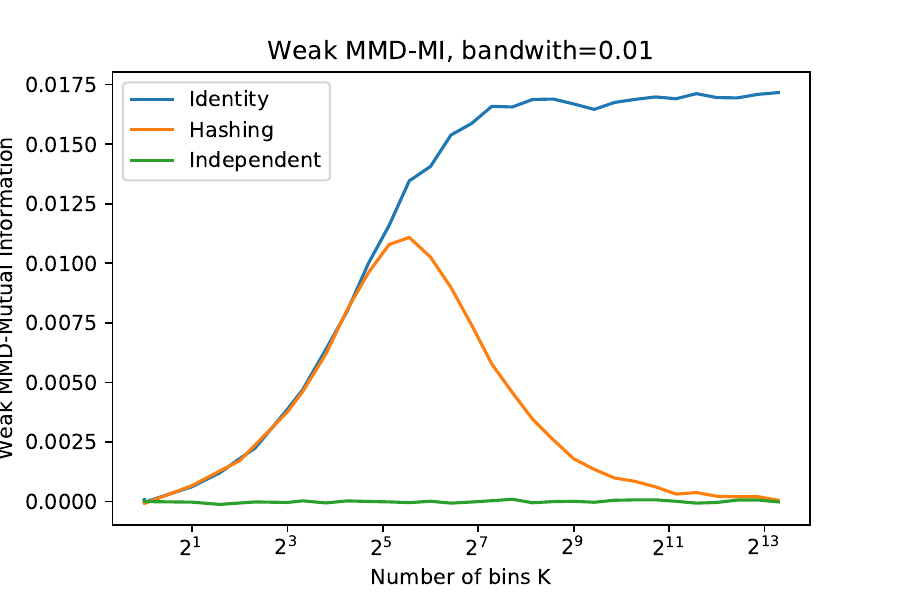}
\includegraphics[width=0.48\textwidth]{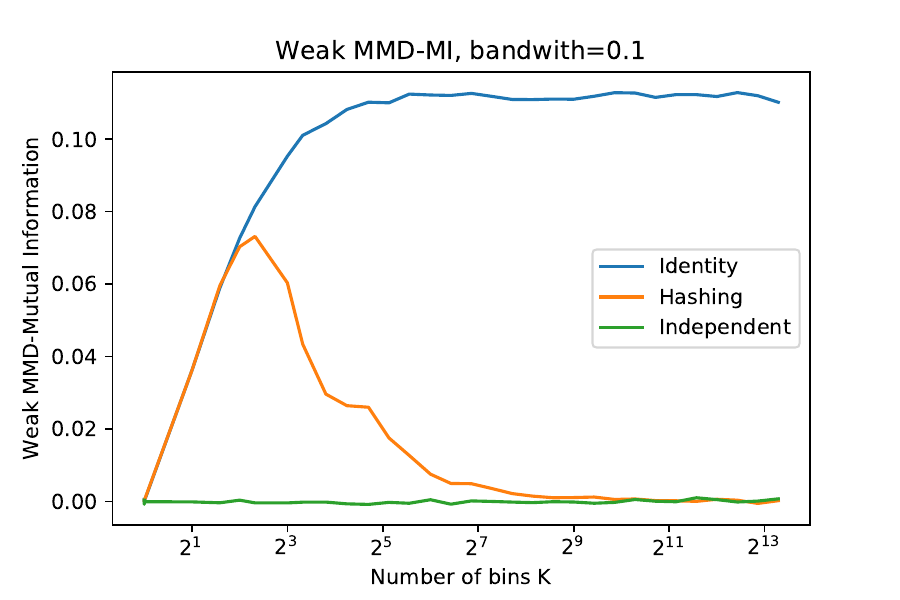}
\includegraphics[width=0.48\textwidth]{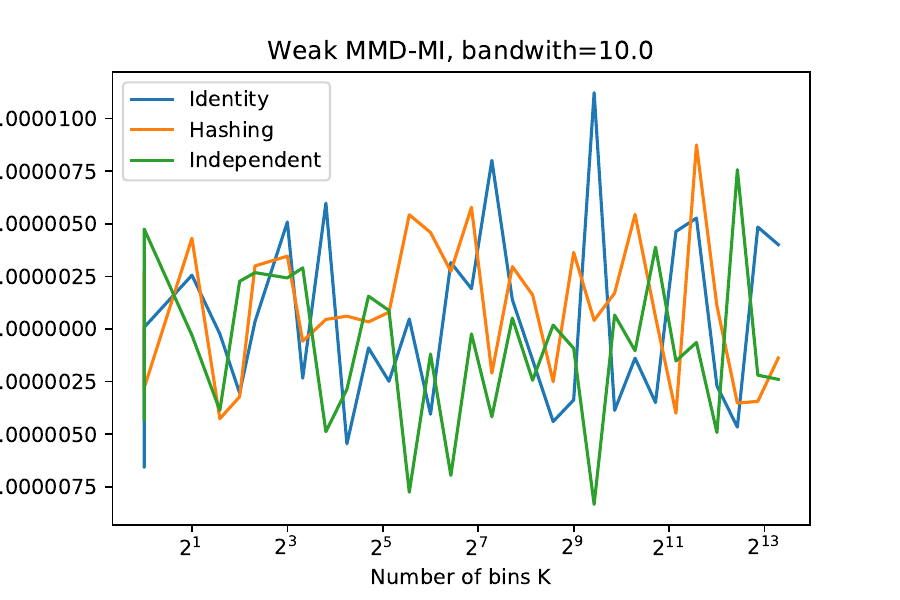}
\caption{Various generalized mutual information (GMI) as a function of the number of bins $K$ for the \textbf{Identity} (blue), \textbf{Hashing} (orange), and \textbf{Independent} (green) distributions. An intuitive notion of mutual information should go to zero for the hashing distribution (orange) as $K$ increases. For smaller values of $K$, MMD-GMI with bandwidths 0.01 and 0.1 increases originally for the hashing distribution, closely following the identity distribution curve (blue), peaks for $K$ roughly equal to half the bandwidth, and then goes to zero to closely follow the independent distribution curve (green). However, when the bandwidth is very large (e.g. 10), all distributions look the same, and MMD-GMI has no discriminative power. \label{fig:hashingresolved}}
\end{figure}
\textbf{[Figure~\ref{fig:hashingresolved}]} For these 3 distributions, we compute the usual KL-based mutual information analytically. We also compute their MMD-GMI with Gaussian kernels of bandwidths $0.01,0.1,10$ using the unbiased U-estimator of $\mathrm{MMD}^2$ proposed in Lemma 6 of~\citet{gretton2007kernel}. It turns out that MMD-GMI with Gaussian kernel and appropriate bandwidth \textit{matches the visual intuition }well. For the hashing distribution, when the number of bins $K$ is small, MMD-GMI increases with $K$ as the bins get smaller and $Y$ looks more determined by $X$, closely tracking the identity distribution curve, until the relationship $Y=h(X)$ becomes too irregular. However, for large enough $K$, the size of each bin becomes smaller than the bandwidth, at which point MMD-GMI goes to zero, and tracks the independent distribution curve.

From the previous result~\eqref{eq:jeqi}, it turns out that when $\sigma_K=hash$, the distribution $p_{X,Y}$ is fit for numerical integration of functions in the Gaussian RKHS (i.e., functions with only low-frequencies). However, it is not the case when $\sigma_K$ is the identity function, and this can be simply shown by considering a function with support far from the diagonal $X=Y$.

\subsection{Semi-Discrete Case\label{sec:semidiscrete}}

When $\mathcal Y$ is discrete with few values (e.g. labels), it makes sense to try to explicitly predict $Y=h(X)$, instead of considering discriminators which take both variables as input. We show that KL-based parametric mutual information (KL-PMI) can be upper (lower) bounded simply by training a classifier to predict $Y$ from $X$ by minimizing the cross-entropy loss, and subsequently subtracting its training (validation) loss from the entropy of $Y$, which is straightforward to estimate for discrete variables with finite values.

\begin{theoremtwo}[Semi-Discrete Mutual Information]
When $Y$ is finite, the mutual information between $X$ and $Y$ can be written as the solution to the following optimization problem :
\begin{align}\label{eq:sdmi}
\mathbf{MI}(X,Y)(X,Y) &= H(Y) - \inf_{T: \mathcal X \times \mathcal Y \to \mathbb R}  \underbrace{\E_{x,y\sim p(x,y)} [-\log q(y|x)]}_{\mathcal L(T)}
\end{align}
where $H(Y)$ is the discrete entropy of the marginal $Y$ and $\mathcal L(T)$ can be reinterpreted as the log-loss or cross-entropy loss between a candidate distribution $q(y|x)$ and $p(y|x)$. Specifically, we define a ``predictor'' $q(y|x) = \frac{p(y) e^{T(x,y)}}{\sum_{y'=1}^{|Y|} p(y') e^{T(x,y')}}$, where $T(x,y)$ can be interpreted as scores for predicting $y$ given $x$, and the inf can be interpreted as finding the score T which maximizes the conditional likelihood for the data from $p$."
\end{theoremtwo}

\begin{proof}
We start from the KL form of mutual information and take conditional expectations using the factorization $p(x,y)=p(x)p(y|x)$ 
\begin{align}
\mathbf{MI}(X,Y) &= \textbf{KL}(p(x,y) || p(x)p(y) ) = \E_{x,y\sim p(x,y)} \left[ \log \left( \frac{p(x,y)}{p(x)p(y)} \right) \right] \\
&= \E_{x\sim p(x)} \E_{y\sim p(y|x)} [\log \frac{p(y|x)}{p(y)}] 
= \E_{x\sim p(x)} \left[ \textbf{KL}(p(y|x) || p(y) ) \right] 
\label{eq:avgkl}
\end{align}
We take the Donsker-Varadhan~\citep{belghazi2018mine} representation of $\textbf{KL}(p(y|x) || p(y) )$ :
\begin{align}
\mathbf{KL}(p(y|x) || p(y) ) = \sup_{M: \mathcal Y \to \mathbb R} \E_{y\sim p(y|x)} [M(y)] - \log \E_{y\sim p(y)} [\exp(M(y))] 
\end{align} 
Injecting this expression in Equation~\eqref{eq:avgkl} yields
\begin{align}
\mathbf{MI}(X,Y) &= \E_{x\sim p(x)} \E_{y\sim p(y|x)} [\log \frac{p(y|x)}{p(y)}] \\
&=  \E_{x\sim p(x)} \left[ \sup_{M: \mathcal Y \to \mathbb R} \left\lbrace \E_{y\sim p(y|x)} [M(y)] - \log \E_{y\sim p(y)} [\exp(M(y))] \right\rbrace  \right] 
\end{align}
The $\sup$ can be taken out of the expectation by considering all functions $T: \mathcal X \times \mathcal Y \to \mathbb R$,
\begin{align}\label{eq:dvstandardmi}
\mathbf{MI}(X,Y) &=  \sup_{T: \mathcal X \times \mathcal Y \to \mathbb R} \E_{x\sim p(x)} \left[  \E_{y\sim p(y|x)} [T(x,y)] - \log \E_{y\sim p(y)} [\exp(T(x,y))] \right] 
\end{align}
Using the fact that $\mathcal Y$ is discrete and finite, we rewrite the expectation in $p(y)$ as a sum :
\begin{align}
\mathbf{MI}(X,Y) &=  \sup_{T: \mathcal X \times \mathcal Y \to \mathbb R} \left\lbrace \E_{x\sim p(x)} \E_{y\sim p(y|x)} \log \frac{e^{T(x,y)}}{\sum_{y'=1}^{|Y|} p(y') e^{T(x,y')}} \right\rbrace
\end{align}
We subtract the marginal entropy $H(Y) = \sum_{y'=1}^{|Y|} p(y') \log p(y')$ which does not depend on~$T$,
\begin{align}
\mathbf{MI}(X,Y)(X,Y) - H(Y) &=\sup_{T: \mathcal X \times \mathcal Y \to \mathbb R}  \E_{x\sim p(x)} \E_{y\sim p(y|x)} \log \underbrace{ \frac{p(y) e^{T(x,y)}}{\sum_{y'=1}^{|Y|} p(y') e^{T(x,y')}}}_{q(y|x)} 
\end{align}
We recognize a weighted softmax over the $T(x,y)$, which we denote $q(y|x)$. After re-arranging, we get :
\begin{align}
\mathbf{MI}(X,Y) &= H(Y) - \inf_{T: \mathcal X \times \mathcal Y \to \mathbb R}  \E_{x,y\sim p(x,y)} [-\log q(y|x)]
\end{align}
\end{proof}
We can modify Equation~\eqref{eq:sdmi} to define semi-discrete parametric mutual informationm by restricting the function $T$ to be in a parametric family $\mathcal T$. 
\begin{definitiontwo}[Semi-Discrete Parametric-KL Mutual Information]
We define the semi-discrete parametric-KL mutual information (KL-SDPMI) as follows:
\begin{align}
\mathbf{SDPMI}(X,Y) &= H(Y) - \inf_{T\in \mathcal T}  \underbrace{\E_{x,y\sim p(x,y)} [-\log q(y|x)]}_{\mathcal L(T)}
\end{align}
where $q(y|x) = \frac{p(y) e^{T(x,y)}}{\sum_{y'=1}^{|Y|} p(y') e^{T(x,y')}}$ and $\mathcal T$ is a parametric family of functions, such as the neural networks with a given architecture. Although it might bee possible to extend the semi-discrete formulation to other divergences, we focus on KL due to its interpretability.
\end{definitiontwo}

Since we recognize a negative log-likelihood loss, we propose the following approach for estimating lower and upper bounds for SDPMI. We independently sample a training set $D_{train}=\lbrace (x_{train}^{(i)}, y_{train}^{(i)}) \sim p_{x,y} \rbrace$  and a validation set $D_{val}=\lbrace (x_{val}^{(i)}, y_{val}^{(i)}) \sim p_{x,y} \rbrace$ from the joint distribution $p_{x,y}$. Then, we train a neural network $q(y|x)$ to predict $y$ from $x$ on the $D_{train}$, and compute the training and validation losses $\mathcal L_{train}, L_{val}$.

\begin{theoremtwo}[KL-SDPMI Bounds]
We can bound the SDPMI using the training and validation SDPMI :
\begin{align}
\underbrace{H(Y) - \E_{D_{train},D_{val}} [\mathcal L_{val}(\widehat T) ]}_{\text{Validation SDPMI}} \leq 
\mathbf{SDPMI}(X,Y) 
\leq
\underbrace{H(Y) - \E_{D_{train}}[\mathcal L_{train}(\widehat T) ] }_{\text{Training SDPMI}}
\end{align}
where $\widehat T\in \mathcal T$ is the minimizer of the training loss $\mathcal{L}_{train}(T) = - \sum_{i=1}^N  \log \frac{p(y_{train}^{(i)}) e^{T(x_{train}^{(i)},y_{train}^{(i)})}}{\sum_{y'=1}^{|Y|} p(y') e^{T(x_{train}^{(i)},y')}}$, $\mathcal L_{val}(\widehat T) =- \sum_{i=1}^{N'}  \log \frac{p(y_{val}^{(i)}) e^{T(x_{val}^{(i)},y_{val}^{(i)})}}{\sum_{y'=1}^{|Y|} p(y') e^{T(x_{val}^{(i)},y')}} $ is the validation loss,
and the expectations are over the sampling of the training and validation sets.
Moreover, the (nonparametric) mutual information can be lower bounded using the validation loss:
\begin{align}
H(Y) - \E_{D_{train},D_{val}} [\mathcal L_{val}(\widehat T) ]\leq \mathbf{SDPMI}(X,Y)\leq \mathbf{MI}(X,Y)
\label{eq:sdpbounds}
\end{align}

\end{theoremtwo}

\begin{proof}
The first inequality comes from the fact that for supervised learning, the training loss (validation loss) is a lower bound (upper bound) of the population loss in expectation. We make use of this inequality to bound the SDPMI in the experiments (Section~\ref{sec:miexp}). 
The second inequality comes from the fact that $\mathcal T$ is a subset of the functions $\X\to\Y$.
\end{proof}

\subsection{GMI for Corrupted Label Datasets (Experiments)\label{sec:miexp}}

We define a range of distributions $q_\alpha$ from less to more corrupted by corrupting the labels of MNIST and SVHN. Specifically, we replace the ground-truth label distribution by taking a mixture between the ground truth (assumed to be deterministic) and uniform distribution over all classes
$$q_\alpha(y|x) = (1-\alpha) p(y|x) + \frac \alpha K$$
where $K=10$ is the number of classes and we take $\alpha\in\lbrace 0,0.25,0.5,0.75,1 \rbrace$. 
For $\alpha=0$ we recover the original distribution with deterministic labels, while for $\alpha=1$ we recover the random-label distribution presented in Section~\ref{sec:permutedmnist}, with labels independent of the image. In practice, we simulate the sampling of the training and test set by corrupting a random fraction $\alpha$ of the ground-truth labels. Because each image $x$ occurs only once in the dataset, it is not obvious that some labels are non-deterministic (corrupted).

\begin{figure}
\centering
\adjustbox{max width=\linewidth}{
\begin{tabular}{l|c|c|c|c}
    \toprule
    Name & Underlying divergence & Function class & Upper Bound & Lower Bound \\
    \midrule
    MI & KL & Nonparametric & Empirical KL  & -\\
    MMD-GMI & MMD & RKHS (Gaussian Kernel) & Empirical MMD & - \\
    Wasserstein-GMI & Wasserstein-1 & $L$-Lipschitz functions & Empirical Wasserstein & - \\
    SDPMI-Linear & KL & Linear & Training SDPMI & Validation SDPMI \\
    SDPMI-RF & KL & Random Forest & Training SDPMI & Validation SDPMI \\
    SDPMI-CNN & KL & CNN & Training SDPMI & Validation SDPMI \\
    \bottomrule
\end{tabular}
}%

\includegraphics[width=\linewidth]{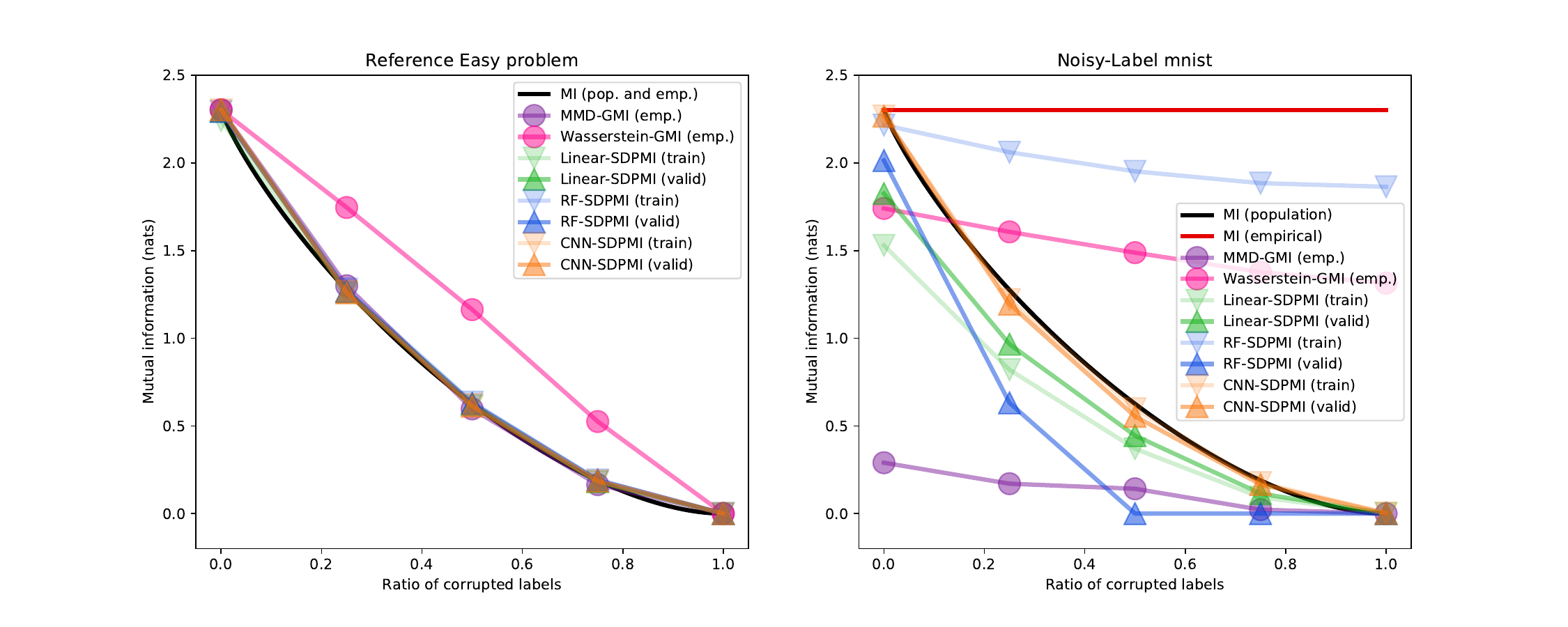}
\includegraphics[width=\linewidth]{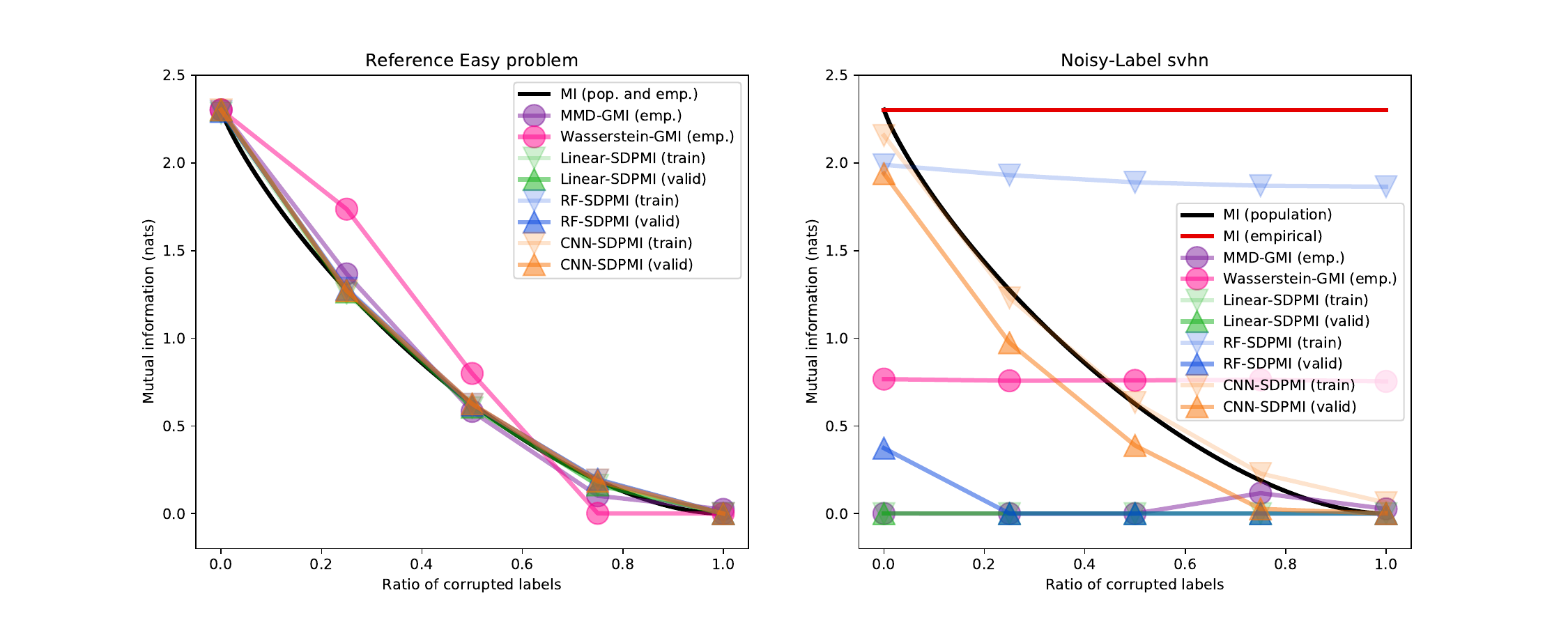}
\caption{\small Estimating various generalized mutual information between image and label on MNIST (\textbf{middle row right column}) and SVHN (\textbf{bottom row right column}) as a function of $\alpha$.
For sanity check and scaling purposes, we also estimate generalized mutual information on reference easy tasks of the same dimensionality (left column pliots).
The mutual information and the semi-discrete parametric mutual information (with linear, random forest and CNN predictors) are expressed in nats.
MMD-GMI and Wasserstein-GMI cannot be directly compared to KL-MI because they have different scales. In order to calibrate the various GMI to each other, we define a \textit{reference} easy task from corrupted MNIST (resp. SVHN), except that the pixels of each flattened image are partitioned into $K$ equal contiguous groups and set to 0 except for the group which index corresponds to the ground-truth label. MMD-GMI and Wasserstein-GMI curves are rescaled such that the maximum value reaches $\log(10)\approx 2.3$ on the reference task.
\label{fig:mimnist}}
\end{figure}

Can we define generalized mutual information which capture the fact that the labels are increasingly corrupted? 
An ideal notion of mutual information should vary \textit{smoothly} and \textit{perceptibly} with respect to $\alpha$.
In \textbf{Figure~\ref{fig:mimnist}}, we compute and plot various GMI as a function of the degree of corruption:
\begin{itemize}
\item \textbf{Mutual Information (Analytical)}. For the specific $q_\alpha$ considered, the true (population) mutual information can be computed analytically and equals 
$ \mathbf{MI}_\alpha(X,Y) = \log K + \log (1 - (1-\frac 1 K)*\alpha) )$, 
which is a smooth function that gradually transitions from $\log K$ to 0, and captures the fact that labels are more and more random. 
\item \textbf{Mutual Information (Empirical)}. On the ``reference easy'' problem (see caption of Figure~\ref{fig:mimnist}), the empirical mutual information approximates the population mutual information closely
because each ``image'' appears many times, since there is only one possible image per class. However, for corrupted MNIST and SVHN, each image appears only once (the odds that the same image appears twice is zero for continuous distributions) and each label appears to be fully determined by the associated image. As a consequence, the empirical mutual information is constant and equal to $\log K$, and fails to capture the fact that labels are increasingly corrupted.
\item \textbf{MMD-GMI (Empirical)}. We compute $\mathbf{MMD}_K(p_{X,Y},p_X\otimes p_Y)$ with kernel $K(x,x',y,y') = \exp(-||x-x'||^2/d) \exp(- 1_{y\neq y'})$ over the training set, where $d$ is the number of dimensions. On the reference tasks, MMD-GMI (purple) varies smoothly with respect to the amount of corruption. However, on corrupted MNIST, MMD-GMI is always close to zero and only slightly sensitive to $\alpha$, which means it considers the label to be mostly independent from the data. This effect is even more obvious for corrupted SVHN. This is consistent with the observation that for generic kernels MMD fails to separate distributions in high-dimensions (Section~\ref{sec:samplecomplexity}).
\item \textbf{Wasserstein-GMI (Empirical)}. Using the Sinkhorn algorithm~\citep{cuturi2013sinkhorn}, we estimate $\mathbf{W}_d(p_{X,Y},p_X\otimes p_Y)$ with base-distance $d((x,y),(x',y'))=||x-x'||/\sqrt d + 1_{y\neq y'}$ over the training set, where $d$ is the number of dimensions. On the reference tasks, Wasserstein-GMI (pink) varies smoothly with respect to the amount of corruption. However, for corrupted MNIST, Wasserstein-GMI is only slightly sensitive to $\alpha$, and seems to be heavily biased, as even the independent case $\alpha=1$ gets assigned a high Wasserstein-GMI. Again, the effect is more obvious for corrupted SVHN, where Wasserstein-GMI does not vary with respect to $\alpha$. This is likely due to the poor sample complexity of the empirical Wasserstein distance, as discussed in Section~\ref{sec:samplecomplexity}.
\item \textbf{KL-SDPMI (Training and Validation)}. We compute the semi-discrete parametric mutual information based on linear (logistic), random-forest, and CNN classifiers, and use Equation~\eqref{eq:sdpbounds} (without the expectations) for estimating upper (training) and lower (validation) bounds for the Semi-Discrete Parametric KL-MI. For the reference tasks, all three Parametric-MI closely approximate the True MI with no apparent gap between training and validation MI. The random-forest Parametric-MI has a large training-validation gap on corrupted MNIST and even more on corrupted SVHN, which indicates severe overfitting. The logistic Parametric-MI varies smoothly with respect to $\alpha$ on corrupted MNIST, althought it slightly underestimates the mutual information (compared to the reference problem). However, on corrupted SVHN, the logistic parametric-MI is not flexible enough to capture the dependence and equals zero for all values of $\alpha$. The CNN Parametric-MI varies gradually with respect to $\alpha$ both for corrupted MNIST and SVHN, with a noticeable but limited gap between train and test-MI. In this case, the CNN Parametric-MI is the most intuitive notion of mutual information.
\end{itemize} 
All divergences are equally informative in the reference easy cases, which act both as a sanity check and a way to calibrate them.
For the harder corrupted MNIST, Logistic and CNN-based SDPMI are the most informative, most likely because they make decent implicit assumptions on the distribution (linear separability, convolutional prior). MMD-GMI and Wasserstein-GMI are not as informative but still somewhat sensitive to $\alpha$, which is intuitive given that they are based pixel-wise metric on the images, which are still a little relevant in the MNIST case. For the hardest noisy-label SVHN case, only the CNN-SDPMI remains informative. The Linear-SPBMI has no sensitivity to $\alpha$ because SVHN is not as linearly separable (as MNIST). MMD-GMI and Wasserstein-GMI fail completely as pixel-wise distances are no longer meaningful for SVHN. 
Random-forest is a terrible estimator for MNIST and SVHN and overfits severely (large training-validation gap) due to its inefficient priors (rule-based classification where each rule is a threshold on a single pixel value).

\section{Related Work} %
\label{sec:related_work}

Closest to our work are the following two papers. \citet{arora2017generalization} argue that analyzing GANs with a nonparametric (optimal discriminator) view does not really make sense, because the usual nonparametric divergences considered have bad sample complexity. They also prove sample complexities for parametric divergences. \citet{liu2017approximation} prove under some conditions that globally minimizing a neural divergence is equivalent to matching all moments that can be represented within the discriminator family. They unify parametric divergences with nonparametric divergences and introduce the notions of strong and weak divergence. However neither of these works focuses on the meaning and practical properties of parametric divergences, as we do here, regarding their suitability for a final task, and paralleling similar questions studied in structured prediction.

Throughout this paper, we have also used several results from the literature to discuss the properties of parametric divergences.
Before the first GAN paper, \citet{sriperumbudur2012empirical} unified traditional Integral Probability Metrics (IPM), analyzed their statistical properties, and proposed to view them as classification problems. Similarly, \citet{reid2011information} show that computing a divergence can be formulated as a classification problem. Later, \citet{nowozin2016f} generalize the GAN objective to any adversarial f-divergence. 
However, the first papers to actually study the effect of restricting the discriminator to be a neural network instead of any function are the \mbox{MMD-GAN} papers: \citet{li2015generative, dziugaite2015training, li2017mmd,mroueh2017mcgan} and \citet{bellemare2017cramer} who give an interpretation of their energy distance framework in terms of moment matching. 
\citet{mohamed2016learning} give many interpretations of generative modeling, including moment-matching, divergence minimization, and density ratio matching.
On the other hand, work has been done to better understand the GAN objective in order to improve its stability~\citep{salimans2016improved}. Subsequently, \citet{arjovsky2017wasserstein} introduce the adversarial Wasserstein distance which makes training much more stable, and \citet{gulrajani2017improved} improve the objective to make it more practical.
Regarding model evaluation, \citet{theis2015note} contains an excellent discussion on the evaluation of generative models, they show in particular that log-likelihood is not a good proxy for the visual quality of samples. \citet{danihelka2017comparison} compare parametric adversarial divergence and likelihood objectives in the special case of RealNVP, a generator with explicit density, and obtain better visual results with the adversarial divergence.  
\citet{belghazi2018mine} propose to use neural networks to estimate mutual information; we propose an alternative formulation in the semi-discrete case and highlight its distinct properties on some toy distributions.
Concerning theoretical understanding of learning in structured prediction, several recent papers are devoted to theoretical understanding of structured prediction such as \citet{cortes2016structured} and \citet{london2016stability} which propose generalization error bounds in the same vein as \citet{osokin2017structured} but with data dependencies.

Our perspective on generative modeling is novel because we ground it on the notion of final task -- what we ultimately care about -- and highlight the multiple reasons why parametric divergences offer a superior framework to define good task losses with respect to a final task; in essence, they provide a more effective and meaningful training signal.
We also perform experiments to determine properties of some parametric divergences, such as invariance/robustness, ability to enforce constraints and properties of interest, as well as the difference with their nonparametric counterparts. To the best of our knowledge, this is the first work that links the task loss generalization error of structured prediction and the adversarial divergences used in generative modeling.

\section{Conclusion}
\label{sec:conclusion}

We have shown that parametric adversarial divergences are not merely lower bounds of nonparametric divergences, but instead have distinct properties which makes them favorable for high dimensional generative modeling.
Among these properties, parametric divergences can scale up to high-dimensional data, and can be tuned to be sensitive to specific aspects of the target distribution.
We have also explored using parametric divergences to define more meaningful notions of mutual information.
An important area of improvement that remains is to compute parametric divergences reliably enough for using them as practical evaluation metrics for generative models.

\acks{This research was partially supported by the Canada Excellence Research Chair in ``Data Science for Real-time
Decision-making'', by the NSERC Discovery Grant RGPIN2017-06936 and by a Google Research Award. These funds were all received and administered by University of Montreal.}

\bibliography{common/biblio}

\newpage

\appendix

In this appendix, we present the following supplementary material:
\begin{itemize}
\item Additional generated samples for the Thin-8 task are in Section~\ref{sec:mil8-more}. 
\item Generated samples for the Sum-25 task are in Section~\ref{sec:morevisual}. 
\end{itemize}

 \section{Experimental results} %
\label{sec:experimental_results}

\FloatBarrier

\subsection{\label{sec:mil8-more}Additional Samples for VAE and GAN}

\begin{figure}[h!]

\centering
\includegraphics[width=0.6\textwidth]{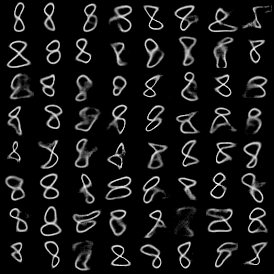}
\includegraphics[width=0.6\textwidth]{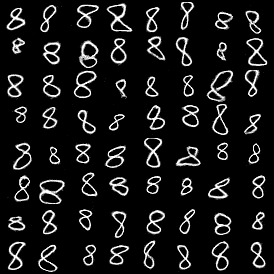}
\caption{VAE (\textbf{top}) and GAN (\textbf{bottom}) samples with 16 latent variables and $32\times 32$ resolution.}
\end{figure}

\begin{figure}[h!]

\centering
\includegraphics[width=0.6\textwidth]{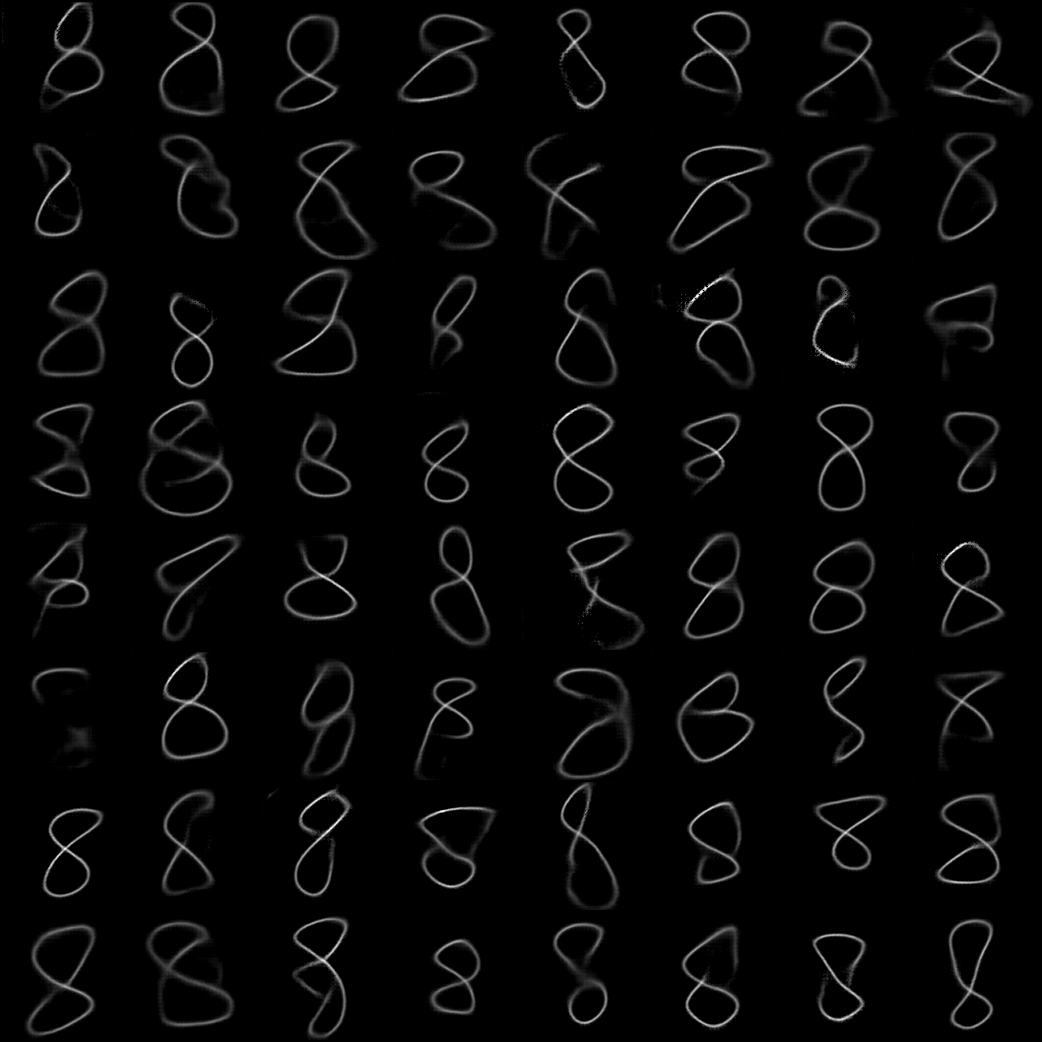}
\includegraphics[width=0.6\textwidth]{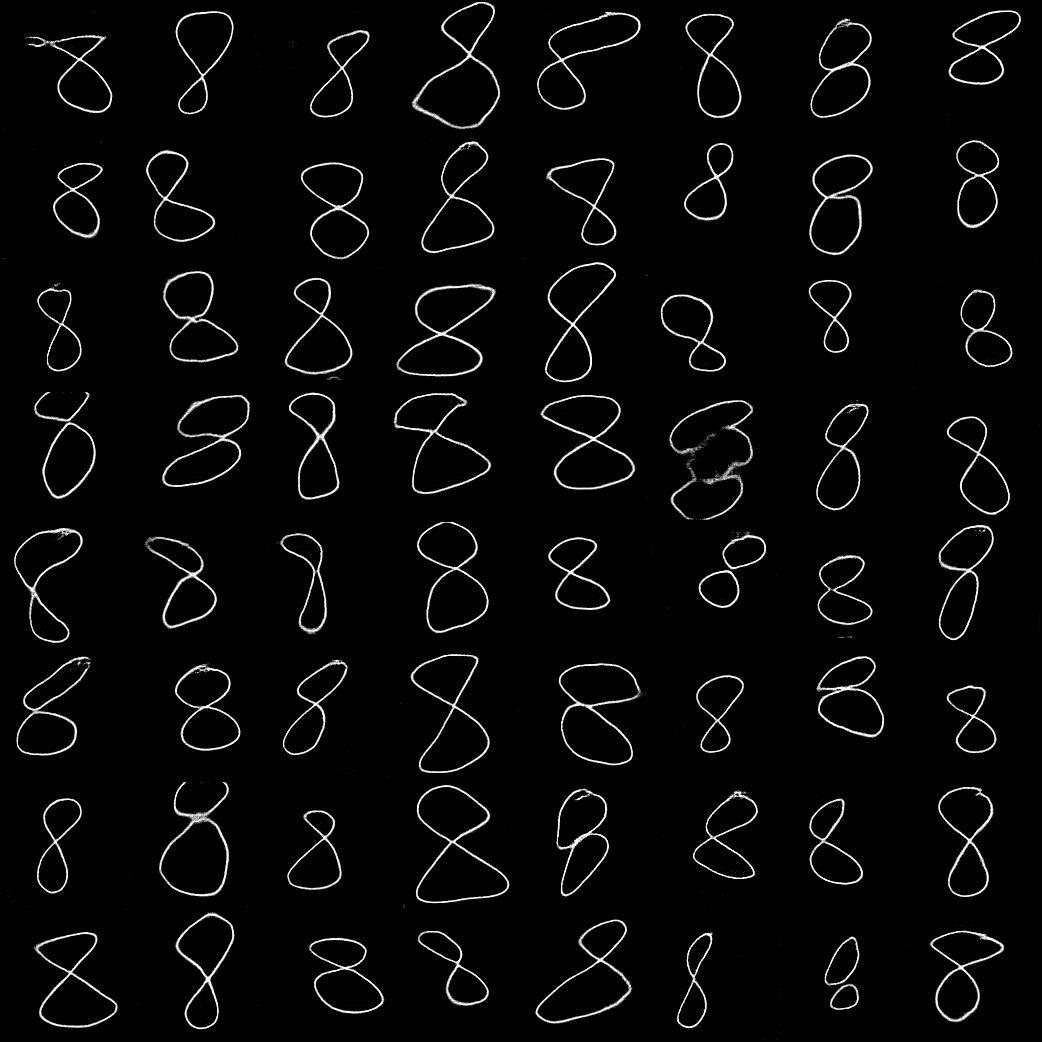}
\caption{VAE (\textbf{top}) and GAN (\textbf{bottom}) samples with 16 latent variables and $128\times 128$ resolution.}
\end{figure}

\begin{figure}[h!]

\centering
\includegraphics[width=0.6\textwidth]{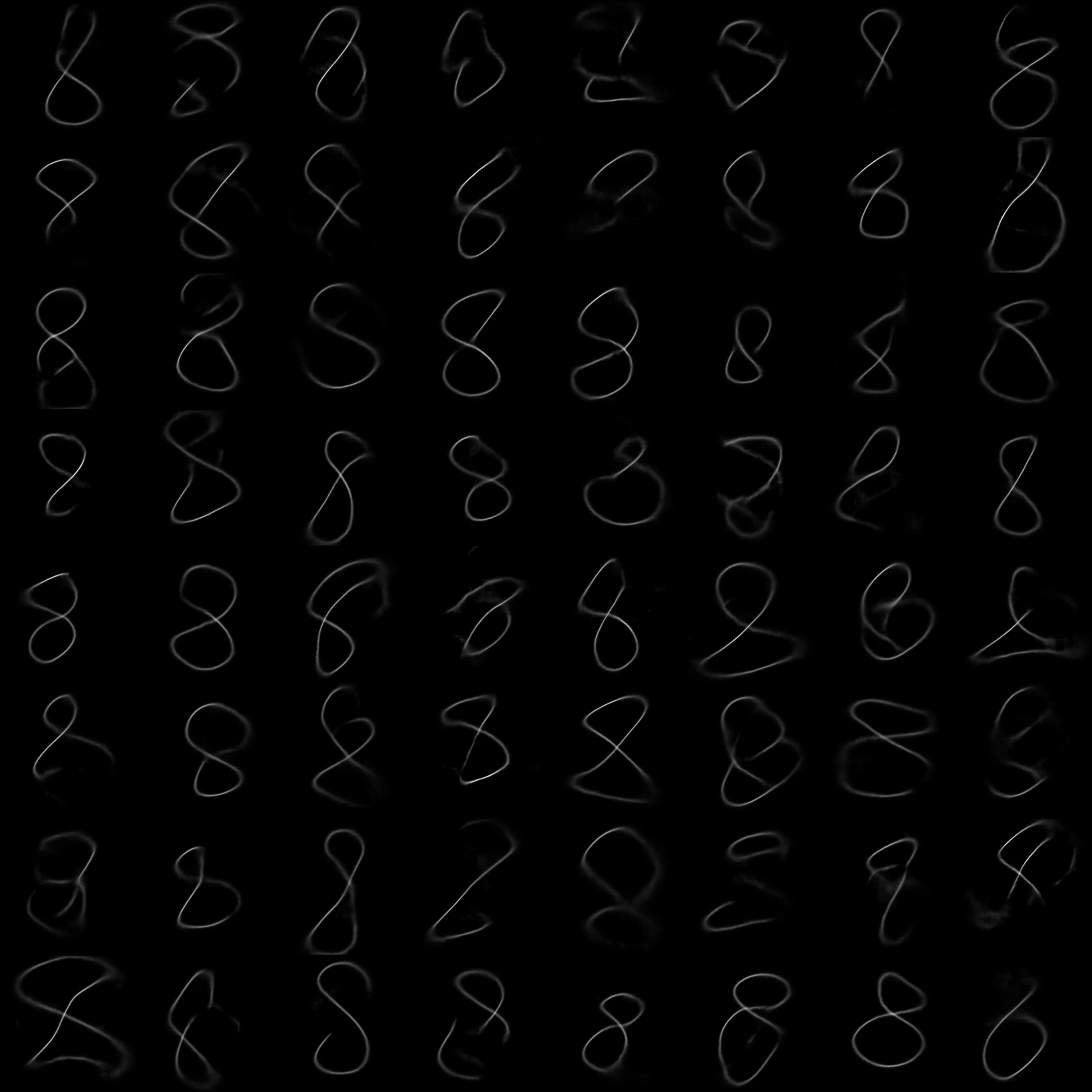}
\includegraphics[width=0.6\textwidth]{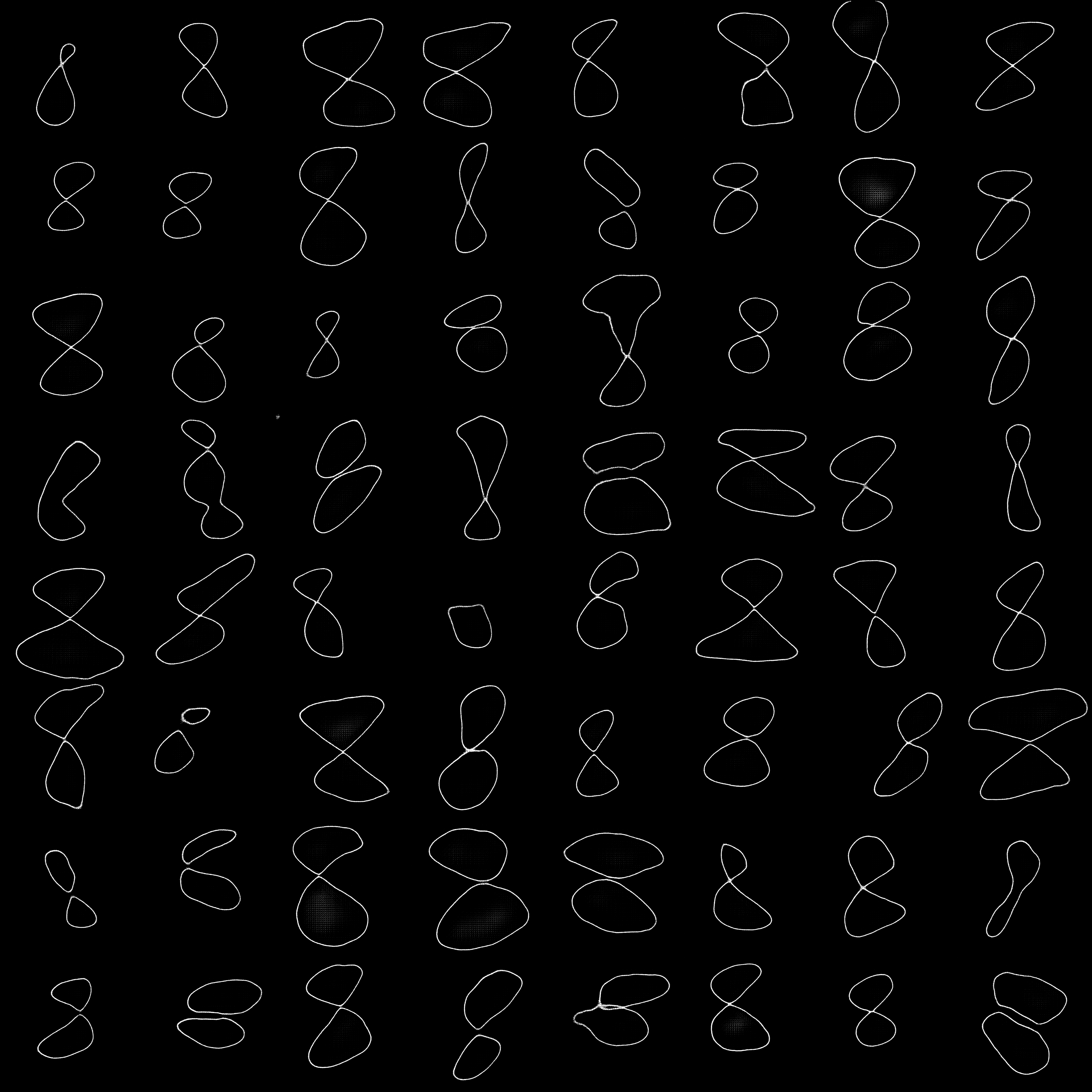}
\caption{VAE (\textbf{top}) and GAN (\textbf{bottom}) samples with 16 latent variables and $512\times 512$ resolution.}
\end{figure}

\clearpage

\FloatBarrier

\subsection{Sum-25: Generated samples\label{sec:morevisual}}

\begin{figure}[h!]

\centering
\includegraphics[width=0.32\textwidth]{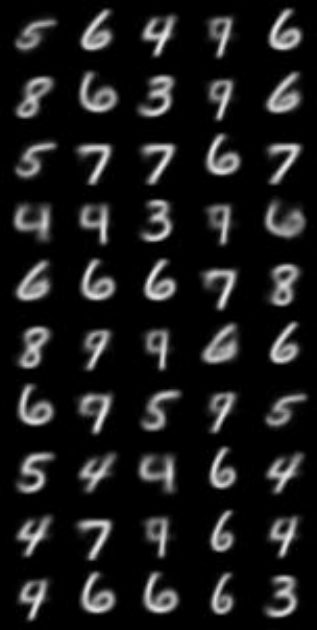}
\includegraphics[width=0.32\textwidth]{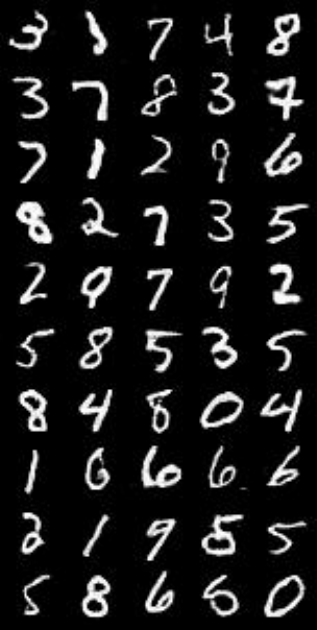}
\includegraphics[width=0.32\textwidth]{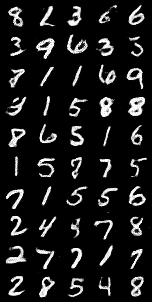}
\caption{Samples from generators with similar architecture and 200 latent variables, trained with the following objectives: VAE (\textbf{left}), WGAN-GP (\textbf{middle}), and WGAN-GP with side-task (\textbf{right}). Each row represents a sample (combination of 5 digits) generated by the model. \label{fig:visual-hyperplane-more}}
\end{figure}

Figure~\ref{fig:visual-hyperplane-more} shows some additional samples from the VAE, WGAN-GP, and WGAN-GP with side-task, trained on the Sum-25. All three generators have 200 latent variables and the same architecture.

\vskip 0.2in

\end{document}